\documentclass[a4paper,fleqn]{cas-sc}

\usepackage[authoryear]{natbib}
\usepackage{enumitem}
\usepackage{subfig}
\usepackage{booktabs}
\usepackage{multirow}
\usepackage{float}
\usepackage{placeins}
\usepackage{algorithm}
\usepackage{algorithmic}
\usepackage{tikz}
\usetikzlibrary{arrows.meta, positioning, shapes.geometric, matrix, calc}

\def\tsc#1{\csdef{#1}{\textsc{\lowercase{#1}}\xspace}}
\tsc{WGM}
\tsc{QE}
\tsc{EP}
\tsc{PMS}
\tsc{BEC}
\tsc{DE}

\begin{document}
\let\WriteBookmarks\relax
\def\floatpagepagefraction{1}
\def\textpagefraction{.001}

\shorttitle{GREAT-EER} %

\shortauthors{A. Lischka, B. Kulcsár}

\title [mode = title]{GREAT-EER: Graph Edge Attention Network for Emergency Evacuation Responses} %

\author[1]{Attila Lischka}[
                        orcid=0009-0003-1371-5996]

\cormark[1]

\ead{lischka@chalmers.se}

\affiliation[1]{organization={Electrical Engineering, Chalmers University of Technology},
    city={Gothenburg},
    country={Sweden}}
\author[1]{Balázs Kulcsár}[orcid=0000-0002-3688-1108]

\cortext[cor1]{Corresponding author}

\begin{abstract}
Emergency situations that require the evacuation of urban areas can arise from man-made causes (e.g., terrorist attacks or industrial accidents) or natural disasters, the latter becoming more frequent due to climate change.
As a result, effective and fast methods to develop evacuation plans are of great importance.
In this work, we identify and propose the Bus Evacuation Orienteering Problem (BEOP), an NP-hard combinatorial optimization problem with the goal of evacuating as many people from an affected area by bus in a short, predefined amount of time. The purpose of bus-based evacuation is to reduce congestion and disorder that arises in purely car-focused evacuation scenarios.
To solve the BEOP, we propose a deep reinforcement learning-based method utilizing graph learning, which, once trained, achieves fast inference speed and is able to create evacuation routes in fractions of seconds. We can bound the gap of our evacuation plans using an MILP formulation.
To validate our method, we create evacuation scenarios for San Francisco using real-world road networks and travel times. We show that we achieve near-optimal solution quality and are further able to investigate how many evacuation vehicles are necessary to achieve certain bus-based evacuation quotas given a predefined evacuation time while keeping run time adequate.
\end{abstract}


\maketitle

\section{Introduction}

In January 2025, a series of wildfires struck the city of Los Angeles, leading to 30 direct fatalities and up to 440 deaths linked to the natural disaster \cite{LAfires}. More than $16,000$ structures where destroyed \cite{LAfires2} and more than $150,000$ people were under evacuation notice \cite{benmarhnia2025beneath}. 
On May 19th 2025, the 300 people living in the village of Blatten, Switzerland had to be evacuated as a precaution \cite{derstandardBlattEvak2025}, before the town was hit and destroyed by a landslide on May 28th of the same year. More than 130 buildings were buried \cite{derstandardBlattWorstCase2025}.
Concurrently, Italian authorities plan for the mass evacuation of the residents near Naples in case the neighboring Plegraean Fields volcano should erupt \cite{guardianCampiFlegrei2023}. 
Investigated evacuation scenarios affect up to $135,000$ people in the proximity of the volcano \cite{charlton2020volcanic}.

Emergency evacuation situations like described above can arise due to a variety of causes, many of them (e.g., landslides, wildfires or inundation) are becoming more frequent due to the effects of the ongoing climate change \cite{fang2025climate, remondo2025increasing, wasserman2023climate, chen2023impacts, yamamoto2021impact}. 
Consequently, the development of effective evacuation strategies is of paramount importance. 

To mitigate the risk of traffic congestion during purely car-based evacuations like it happened during Hurricane Irma in 2017 \cite{FENG2021102788} when highways all over Florida were blocked for days, bus-based, collective evacuation strategies could be of advantage. Bus-based evacuation could also reduce the risk of power-shortages due to grid-limitations that could arise during car-based evacuations using electric vehicles \cite{FENG2020102458}.

Existing works, e.g., dealing with the \textit{Bus Evacuation Problem} focus on evacuating a given set of people in as little time as possible \cite{feng2024approximation}. 
However, in reality, it might not be possible reach every evacuation point in an evacuation plan since the available evacuation time, capacity and resources in practice might be limited. 
Consequently, the objective shifts to reaching as many evacuation points under a set of constraints as possible, while assuming that the remainder of the people will evacuate by car themselves. 
The problem thus becomes related to the capacitated team orienteering problem (CTOP). The CTOP is an NP-hard combinatorial optimization problem with the goal of collecting as many prizes in a given amount of time as possible using several cooperating vehicles of limited carrying capacity and consecutive return to a designated ``depot'' location. Compared to the conventional definition of the CTOP where vehicles must not leave the depot again once they arrive, we change the problem to allow multiple departures and arrivals of each vehicle at the depot node.
Further, we consider additional travel constraints that result in limited time to visit certain points in our problem instance. In a way, we investigate a multi-tour capacitated team orienteering problem with time windows. Since in essence our problem is related to both the BEP and the CTOP, we name it the Bus Evacuation Orienteering Problem (BEOP). 
In the context of our evacuation setting the problem looks like this:
All evacuation operations are centered around a safe location where people can find shelter. To bring people to the shelter, we have several evacuation vehicles with a known capacity (e.g., three buses with 50 seats each). We have a predetermined time (e.g. 2 hours) to evacuate as many people as possible and bring them to the safe location. When a bus has filled up all its seats, it returns to the safe location and drops the evacuees off. If there is time left, it leaves again and evacuates further people. 
The people are collected from evacuation points (e.g., their houses) with a known amount of people to evacuate.
It is possible, however, that some evacuees are not willing to wait until the end of the evacuation time for a pickup but instead want to be picked up during, e.g., the first half of the evacuation period, resulting in an evacuation time window. If the bus does not arrive at such an evacuation point before the time window closes, the evacuees of this point will travel to the shelter by car themselves.
In general, we assume knowledge of the travel time between two different evacuation points and the return time to the depot.
Further, to not endanger any first responders and emergency rescue personal, we need all rescue vehicles to be back at the safe location before the maximum evacuation time is exceeded.
We assume that the remainder of the people that did not get evacuated by the busses go to a safe location on their own by car. However, to reduce the risk of congestion and disorder during a purely car-based evacuation, we want to evacuate as many people by bus as possible.

\subsection{Literature Review}
\subsubsection{Machine Learning-Based Methods to Solve Orienteering Problems}
In recent years, several machine learning-based methods to tackle orienteering problems have been proposed. 
One of the first papers to solve Euclidean OP using deep learning is the attention model (AM) by \cite{kool2018attention}. \cite{xin2020multidecoderattentionmodelembedding} proposes a multi-decoder AM to generate a more diverse solution set and by this increases performance. \cite{kim2023symncoleveragingsymmetricityneural} exploit symmetries in the solution space (TSP) and representation space (EUC COP) for better learning of different VRPs, among them OP.
\cite{lischka2025greatarchitectureedgebasedgraph} propose a novel deep graph-learning-based methods named GREAT to tackle asymmetric and symmetric OP. \cite{drakulic2023bqncobisimulationquotientingefficient} use a supervised-learning-based (SL) method that can be trained on small instances of different VRPs and later on generalizes well to larger instances. By this, they are able to solve Euclidean OP for up to 1000 nodes. In \cite{drakulic2025goalgeneralistcombinatorialoptimization}, this SL framework is generalized to solve different routing problems (among them OP) at once. This shows that trained deep networks can generalize between routing problems and sets the path for the development of a ``VRP-foundation model''.
\cite{yao2024rethinking} proposes another SL based methods to train deep learning-based routing models (among them for OP). In particular, they investigate data augmentation methods to drastically reduce the amount of labelled ground truth data needed for training. By using this SL approach, they can reduce the training time needed to achieve similar inference performance with RL approaches significantly.

\cite{gama2021reinforcementlearningapproachorienteering} investigates the more restrictive OP with time windows (OPTW) where each node can only be visited between a specified start and end time. They solve this problem using a pointer network trained with reinforcement learning and relate OPTW to the tourist trip design problem. 
\cite{li2024reinforcementlearningapproachesorienteering} uses value function approximation (VFA) to learn a stochastic variant of OP for parcel delivery. In particular, it unknown when a parcel is ready for delivery at the depot node. The aim is to deliver as many parcels to customers as possible before a deadline.

We refer the interested reader to \cite{ZHOU2025104278} for an extensive overview on how machine learning has been used to tackle routing problems (among them OP).

\subsubsection{Capacitated Orienteering Problem}
An overview over different OP extensions, among them the capacitated team orienteering problem or variants with stochastic rewards and travel times can be found in \cite{vansteenwegen2019other}.
The capacitated orienteering problem (CTOP) was first studied in \cite{Archetti01062009}. 
Compared to the simple OP problem, where the goal is to maximize the amount of collected rewards while respecting a maximum travel distance or time, in the CTOP, a whole fleet of vehicles work together to maximize the amount of maximized rewards. Moreover, each prize also has an associated demand and a vehicle can only serve a certain capacity of demands. After a vehicle's demands or travel time are exhausted, it needs to go back to its starting point.
As pointed out in \cite{gunawan2016orienteering}, the CTOP can be solved using search algorithms \cite{tarantilis2013capacitated} or branch-and-prize methods \cite{archetti2013optimal}.
Compared to these deterministic works, \cite{shiri2024capacitated} investigates a stochastic variant of the online CTOP by developing online optimization algorithms. In particular, the exact prize and demand of a node in the considered OP instances are unknown.

\subsubsection{Bus Evacuation Problem}
The Bus Evacuation Problem (BEP) describes a set of Vehicle Routing Problem (VRP) related combinatorial optimization scenarios \cite{bish2011planning}.
In essence, solving the BEP breaks down to route a set of buses in a way to collect all people in an predefined area.
The original BEP by \cite{bish2011planning} included yard nodes (where buses start their tours), demand nodes (locations where evacuees are waiting) and shelter nodes (safe locations the evacuees are being brought to). The buses and shelter nodes have associated capacities that must be respected and the objective is to minimize the evacuation time. 
Consequently, the BEP is related to the pickup and delivery problem.
Additionally, for the BEP, it can be assumed that individual pick-up locations can have a higher demand than a buses capacity, requiring multiple bus visits and resulting in a Split Delivery VRP related problem.
A selection of works dealing with the BEP are \cite{goerigk2013branch, dikas2016solving, feng2024approximation, ZHAO2020285}, ranging from branching-based methods to approximation algorithms.
We refer the interested reader to \cite{feng2024approximation} for a small overview of different works tackling variants of BEP using different algorithm families.

In an extension of the BEP in \cite{goerigk2014combining}, the selection of the shelter and gathering locations are optimized together with the bus routing. In \cite{GOERIGK201482}, individual traffic is additionally also considered and \cite{GOERIGK201566} develops a method incorporating uncertainty.
In general all variants of the BEP have in common, however, that their objective is to minimize the overall evacuation time or travel distance when serving all evacuation points. 
This is in contrast to OP settings, where it is possible that there is not enough time available to visit every node and the aim is to visit as many nodes as possible.
In \cite{sayyady2010optimizing} a variant of the BEP is studied that minimizes travel time while at the same time minimizing casualties. This is done by allowing the assignment of evacuees to ``sink nodes'' (corresponding to no evacuation) which are associated with very high travel costs. Therefore, when minimizing travel time, the model implicitly also aims to minimize the number of casualties.

A work that combines reinforcement learning and bus based evacuation is \cite{tang2025strategizing}. There, public transport is rerouted to bring people to safe shelters. Their reward function aims to minimize evacuation and waiting times while penalizing inequities during the evacuation.

\subsubsection{OP for Emergency Evacuation}
The study by \cite{baffo2017orienteering} investigates multi-origin CTOP to evacuate people in emergency situations from industrial plants. To solve the problem, an ant colony optimization (ACO) is proposed and evaluated on 2 instances with different amounts of vehicles and origin combinations. The solutions are benchmarked against an exact solver and it is shown that the ACO achieves near-optimal solutions. \cite{DOLINSKAYA20181} explores adaptive orienteering problem with stochastic travel times in the context of search and rescues in post-disaster-settings.

\subsection{Contributions}

Our paper contains the following main contributions:

\begin{itemize}
    \item Model: introduction of the BEOP as a critical component in emergency based evacuations using buses.
    \begin{itemize}
        \item Introduction of a mixed integer linear program formulation of the BEOP
        \item Introduction of a Markov Decision Process for the BEOP
    \end{itemize}
    \item Method: a deep graph reinforcement learning framework to solve the BEOP.
    \item A real-world experimental evaluation
    \begin{itemize}
        \item In a deterministic setting, our trained RL model allows us to create high-quality evacuation plans a priory before an emergency situation arises.
        \item Our trained model gives guidance on how many evacuation vehicles are needed in an emergency scenario given parameters like evacuation time or vehicle capacity to achieve certain evacuation quotas.
        \item In a stochastic setting, our trained RL model can react to stochastic changes resulting from the emergency situation and adjust its routes accordingly.
    \end{itemize}
\end{itemize}

First, we introduce the BEOP as a key component of bus-based evacuation in emergency settings. To this aim, we introduce both, a mixed-integer linear program (MILP) describing the problem and a Markov Decision Process (MDP) that can model the setting.

\begin{figure}
    \centering
    \includegraphics[width=0.9\linewidth]{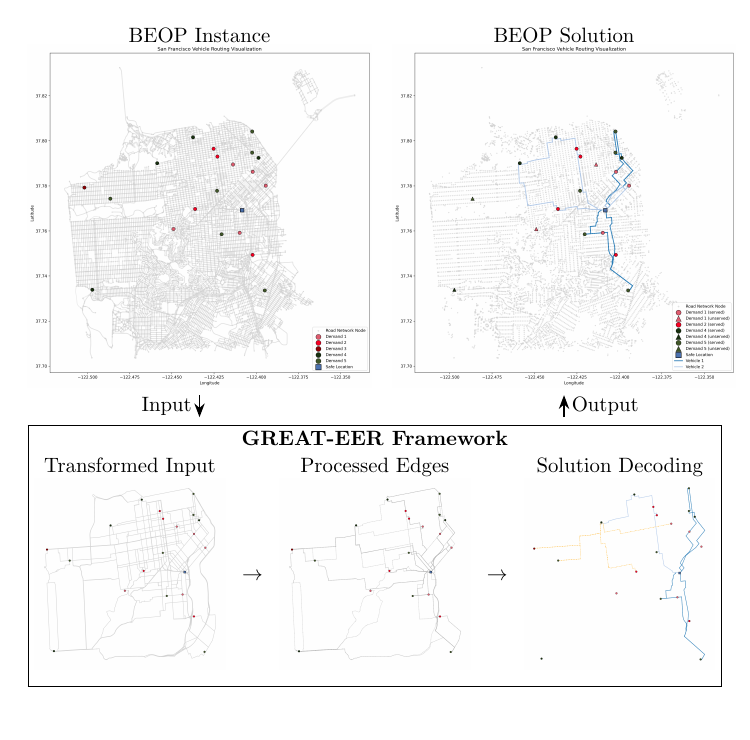}
    \caption{GREAT-EER Framework for the BEOP: a BEOP instance is transformed into a graph that encodes the pairwise shortest paths between all relevant nodes. Then, the GREAT encoder is used to process this edge-based graph. Finally, a pointer network iteratively selects nodes to visit and their order. The resulting tour is returned as the output.}
    \label{fig:greateer-framework}
\end{figure}

Second, to solve the BEOP, we develop a graph-learning-based deep reinforcement learning framework. The framework uses the graph edge attention network (GREAT) as an encoder architecture and a multi-pointer network as the decoding part that iteratively selects the next graph node to visit for evacuation (Figure \ref{fig:greateer-framework}). The framework can be trained in an unsupervised setting using reinforcement learning and eliminating the need for an oracle solver. We note that our framework is trained offline once and, after being trained, it can solve any BEOP instance of similar characteristics fast, achieving high solution quality.

Third, we confirm and evaluate the proposed solution method by creating an experimental setting based on real world data from San Francisco. We use OpenStreetMap to obtain realistic travel times between points of interests and a publicly available Uber dataset to infer which locations in the city are frequented by people and, therefore, interesting evacuation points.
We compare the solution quality to the output of an MILP-based Gurobi model and, by this, can bound its performance and evaluate its optimality gap. Further, we compare our learning-based algorithm's performance to a simple custom developed greedy algorithm. 
We use our trained model to investigate how many buses of certain capacities are requires to achieve a predefined bus-evacuation-quota (e.g. 60\%) in a given amount of time. Additionally, we demonstrate that our trained model achieves strong performance on data different from the training data by considering larger BEOP instances during testing, as well as instances that contain ``hazard zones'' that include blocked roads. Moreover, we show that our model also works in a stochastic online setting, by reacting to deviations from the expected travel times and evacuation demands and adjusting its routing decisions accordingly.

The remainder of this article is structured as follows: We introduce and define the BEOP in Section \ref{sec:problem} providing a MILP formulation and a MDP describing the setting. Moreover, we show that BEOP is NP-hard. In Section \ref{sec:framework} we describe the graph neural network-based architecture that captures a BEOP instance and is used to construct the evacuation routes. Further, we introduce the reinforcement learning framework used to train the neural model. The experimental evaluation in a deterministic planned way and in a stochastic emergency setting can be found in section \ref{sec:experiments}. The experimental setting is created from publicly available data of San Francisco. Finally, we conclude in \ref{sec:conclusion}.

\section{Problem Formulation: Bus Evacuation Orienteering Problem} \label{sec:problem}

We define the Bus Evacuation Orienteering Problem (BEOP) as a graph problem:
Let $V = \{0, 1, \dots, n\}$ be a set of $n+1$ graph nodes where $0$ corresponds to the safe depot where rescue vehicles start and end their tours and $\{1, 2, \dots, n\}$ are the $n$ evacuation locations we can pickup people from. 
Further, we have a set of directed edges $E = \{(u,v) \mid u, v \in V, u \neq v\}$ with associated travel times $t_{uv}, u,v \in V$ where in general it holds that $t_{uv} \neq t_{vu}$, i.e. the travel times are asymmetric.
Further, for all $v \in V$ we have an associated prize $p_v$ that corresponds to the amount of people that are to be rescued from location $v$. Similarly, we have a demand $d_v$ for all $v \in V$ which corresponds to the amount of space the rescue vehicle rescuing the people at node $v$ will occupy. 
In our setting, we generally assume that $d_v = p_v$ for all nodes. However, we note that in reality this does not need to hold since some individuals might need more space than others or their associated prize is different (e.g., favoring of the evacuation of children).

For the evacuation, we have $k$ rescue vehicles (i.e. buses), each with a capacity of $C_k$ that corresponds to the maximum amount of people it can carry at once.
For the sake of simplicity we assume that all vehicles have the same capacity in our evacuation scenarios, i.e. $C_k = C$ for all $k$. Due to the capacity constraint, a vehicle cannot visit a subset of nodes $U \subseteq V$ with $\sum_{u in U} d_u > C$ without visiting the depot in between. 
Let $\tau_k = (i_{k_0}, i_{k_1}, \dots, i_{k_m})$ be the tour executed by vehicle $k$. 
Then it most hold that $i_{k_0} = i_{k_m} = 0$ which means that the tour starts and ends in the safe depot.
We explicitly emphasize that a vehicle is allowed to return to the depot intermediately (i.e., for some $j \neq 0, m$ it holds that $i_{k_j} = 0$) to deliver passengers and reset its capacity. This twist makes the problem different from CTOP where each vehicle can only execute a single tour.
This iterative arrival and departure at the safe location is fundamental to an emergency evacuation, since it would not make sense for a vehicle to not resume evacuations as long as there is still further time left.
For the different evacuation tours of different vehicles $k, k', k\neq k'$ it holds that $\{i | i \in \tau_k\} \cap \{i | i \in \tau_{k'}\} =  \{ 0 \}$. This means no node apart from the depot is visited by more than one vehicle. 
Moreover, for a vehicle $k$ it holds that $i_{k_j} = i_{k_{j'}}, j \neq j' \rightarrow i_{k_j} = 0$ which means that no node other than the depot can be visited more than once by a vehicle $k$ itself either.

Due to the orienteering problem nature of the BEOP, we also have a maximum evacuation time $T$. Consequently, for every vehicle $k$ it must hold that its final return to the depot is before the evacuation time $T$ is over, i.e. $\sum_{j=0}^{m-1} t_{i_{k_j}, i_{k_{j+1}}} \leq T$. We note that the evacuation time of a vehicle is computed over all its subtours, which is why it is not possible to just think of BEOP as a CTOP with additional vehicles that only do one subtour each.
Furthermore, it is possible for some nodes $u \in V$ to have individual ``deadlines'' for the evacuation, i.e. the evacuees of node $u$ want to be picked up before time $f_u$ after the evacuation started.
this means, that if $i_{k_\ell} = u$ for some vehicle $k$, then $\sum_{j=0}^{\ell-1} t_{i_{k_j}, i_{k_{j+1}}} \leq f_u$.

\newtheorem{theorem}{Theorem}
\begin{theorem}
The Bus Evacuation Orienteering Problem with satisfied triangle-inequality for the travel times is NP-hard.
\label{thm}
\end{theorem}

\newproof{pot}{Proof of Theorem \ref{thm}}
\begin{pot}
It is known that basic OP is NP-hard \cite{vansteenwegen2019orienteering}. 
As a result, we can show that BEOP is NP-hard by reduction, i.e., if we can solve BEOP efficiently in polynomial time we can also solve OP efficiently in polynomial time by transforming (\textit{reducing}) one problem into another.
Note that we can transform an OP instance into a BEOP instance by keeping the same maximum travel time and the same prizes for each node. Further, the BEOP instance has one vehicle only. The vehicle's capacity is set to $n$ (for an OP instance with $n$ nodes) and each node's demand is set to $1$.
It is easy to see that we can perform this transformation in polynomial time in the number of nodes $n$.
When solving this BEOP instance, it is possible that the bus in the BEOP solution has an intermediate return to the depot and leaves again (since BEOP allows subtours). Let $(i, 0, j)$ be such a subsequence in the solution tour that visits the depot and afterwards leaves. 
We observe that the capacity of the bus is big enough to visit all customers without intermediate return to the depot. As a result, we can skip the depot and directly go from $i$ to $j$. Since the triangle inequality holds, we know that $t_{i0} + t_{0j} \geq t_{ij}$ and, therefore, that skipping node $0$ in the solution also respects the maximum travel time. 
Such unnecessary depot visits in the BEOP are possible because it is possible that there is more than one optimal solution and the time saved by not going to the depot does not suffice to visit any additional other nodes (or because there is enough time to visit all nodes of the instance either way).
We emphasize that real-world graphs are typically asymmetric (considering travel times) but respect the triangle inequality \footnote{Consider a matrix $T$ that contains the pair-wise shortest travel times between all nodes in a road-network (i.e., length of the shortest paths in the network w.r.t. travel time). Assume there exist three nodes $i,j,k$ in the road network such that $T_{ij} + T_{jk} > T_{ik}$. Then there is a faster path from $i$ to $k$ than what is reported in the matrix - a contradiction to the matrix containing the pair-wise shortest travel times.}.
Deleting all unnecessary depot visits of a BEOP solution can be achieved in polynomial time since the size of the solution tour of the BEOP is linear in $n$.
Therefore, if a polynomial time algorithm for the BEOP exists, we can also solve OP in polynomial time since the transformation from an OP instance to a BEOP instance and the subsequent recovery of an OP solution from the BEOP solution can be achieved in polynomial time.
\end{pot}

\newdefinition{rmk}{Remark}
\begin{rmk}
It is straightforward to see that the decision variant of the BEOP (``Is there a feasible route that evacuates at least X people?'') is verifiable in polynomial time. Therefore, BEOP is NP-complete.
\end{rmk}

\subsection{Optimization Program for the BEOP}
Below, we provide a mixed integer linear program (MILP) formulation for the BEOP based on \cite{shiri2024capacitated} for simple CTOP and adjusted to allow multiple vehicle subtours. We note that the BEOP (compared to the CTOP) also has ``time windows'' for some nodes which means that an evacuation of a node is, e.g., possible during the first half of the evacuation period only. The constraints for the time window is taken from the team orienteering problem with time windows formulation in \cite{vansteenwegen2019orienteering} and has again been adjusted for subtours. 
\begin{align}
    \max &\sum_{k^* \in \textcolor{black}{K^*}} \sum_{i \in V \setminus \{0\}} p_i y_i^{k^*} \\
    \text{s.t.} & \sum_{i \in V \setminus \{0\}} d_i y_i^{k^*} \leq C, \quad \forall k^* \in \textcolor{black}{K^*} \\
    & \textcolor{black}{\sum_{k^* \in \left[ k \right]}}\sum_{i \in V} \sum_{j \in V \setminus \{i\}} t_{ij} x_{ij}^{\textcolor{black}{k^*}} \leq T, \quad \forall k \in K \\
    & \sum_{i \in V, i \ne l} x_{il}^{k^*} = \sum_{j \in V, j \ne l} x_{lj}^{k^*}, \quad  \forall l \in V, \forall k^* \in \textcolor{black}{K^*} \\
    & \sum_{i \in V} x_{0i}^{k^*} \textcolor{black}{\leq} 1, \quad \forall k^* \in \textcolor{black}{K^*} \\
    & \sum_{i \in V \setminus \{ j\}} x_{ij}^{k^*} = y_j^{k^*}, \quad \forall k^* \in \textcolor{black}{K^*}, \forall j \in V \setminus \{0\} \\
    & \sum_{k^*\in K^*} y_j^{k^*} \leq 1, \quad  \forall j \in V \\
    & u_j^{k^*} \geq u_i^{k^*} + 1 - |V| \times (1 - x_{ij}^{k^*}), \quad \forall i, j \in V, i \neq j, j \neq 0, \forall k^* \in \textcolor{black}{K^*}
    \\
    & u_i^{k^*} \leq \sum_{j \in V \setminus \{0\}}y_j^{k^*}, \quad \forall k^* \in \textcolor{black}{K^*}, \forall i \in V \setminus \{0\} \\
    & u_0^{k^*} = 0, \quad  \forall k^* \in \textcolor{black}{K^*}
    \\
    & s_i^{k^*} \leq f_i \quad \forall k^* \in K^*, \forall i \in V \\
    & s_j^{k^*} \geq s_i^{k^*} + t_{ij} - M_T \times (1-x_{ij}^{k^*}) \quad \forall k^* \in K^*, \forall i \in V, \forall j \in V \setminus \{0\}, i \neq j \\
    & s_0^{k_1} = 0 \quad \forall k \in K \\
    & s_0^{k_n} = \sum_{i \in V} \sum_{j \in V \setminus \{i\}} \sum_{m < n} x_{ij}^{k_m} t_{ij} \quad \forall k_n \in \left[ k \right] \setminus \{k_1\} \\
    & s_i^{k^*} \geq 0, \quad  \forall i \in V, \forall k^* \in \textcolor{black}{K^*}
    \\
    & u_i^{k^*} \geq 0, \quad  \forall i \in V, \forall k^* \in \textcolor{black}{K^*}
    \\
    & y_i^{k^*} \in \{0, 1\}, \quad \forall k^* \in \textcolor{black}{K^*}, \forall i \in V \setminus \{0\} \\
    & x_{ij}^{k^*} \in \{0, 1\}, \quad \forall i, j \in V, i \neq j, \forall k^* \in \textcolor{black}{K^*}
\end{align}
Our problem contains four types of variables: $x_{ij}^{k^*}$ (18) which is binary and indicates whether vehicle subtour $k^*$ traverses the edge from node $i$ to node $j$, $y_i^{k^*}$ (17) which is also binary and indicates that node $i$ is visited by vehicle subtour $k^*$ and $u_i^{k^*}$ (16) which is continuous and indicates the order in which vehicle subtour $k$ visits a node $i$ and $s_i^{k^*}$ (15) which is also continuous and indicates the arrival time of vehicle k to node $i$. If the BEOP does not contain time windows, we do not need (11) - (15).

The purpose of the individual terms is as follows:
(1) is the objective function which maximizes the collected prizes of the BEOP, i.e., the amount of people evacuated.
(2) ensures that each vehicle subtour respects the vehicle's capacity. (3) ensures that all sum of the travel times of all subtours a single vehicle does, does not exceed the maximum evacuation time. (4) corresponds to the flow conservation and guarantees that if a vehicle subtour $k$ leads to node $l$, it also leaves the node again. (5) ensures that each subtour leaves the depot at most once. (6) links the binary edge variables with the binary node variables and ensures that if a vehicle subtour $k$ traverses and edge $x_{ij}^k$ node $j$'s variable is set to visited. (7) certifies that each node is visited at most once. (8-10) ensures there are no subtours within a subtour. (11-14) ensures the arrival time at each evacuation node respects the latest possible arrival time for each node.

Note that the set $K^*$ is the set of all vehicle subtours, whereas the set $K$ is the set of all vehicles. Further, $\left[k \right]$ is the sequence (which is ordered) of all subtours a vehicle $k \in K$ does.
In theory, the optimal solution of a BEOP instance could have a very large amount of vehicle subtours for a single vehicle (if a single vehicle visits each node in an individual subtour, we would have $|V \setminus \{0\}|$ subtours). Due to practicality reasons, we restrict the amount of subtours a single vehicle can do, to e.g. three. By this we get:
$\left[k \right] = (k_1, k_2, k_3)$ and $ K^* = \bigcup_{k \in K} \text{set}(\left[k \right] )$. 

While this MILP formulation cannot directly be used to solve stochastic variants of the BEOP, it can be extended by modifying (1) to maximize expected prize. Similarly, we can adapt (2) and (3) to incorporate stochastic demands and travel times through appropriate stochastic constraints. As is, it can serve as an oracle baseline by passing the model the realized values of the stochastic parameters.

\begin{table}[width=.9\linewidth,cols=2,pos=h]
\caption{Notation for BEOP}\label{tbl1}
\begin{tabular*}{\tblwidth}{@{} LLLL@{} }
\toprule
Symbol & Description \\
\midrule
\textit{General problem properties} \\
$n$ & number of evacuation nodes  \\
$V = \{0, 1, \dots, n\}$ & set of safe location (or ``depot'') $0$ and evacuation nodes \\
$E = \{(u,v) \mid u,v \in V, u \neq v\}$ & set of directed edges  \\
$T$ & maximum evacuation time  \\
$K = \{1, \dots, k\}$ & set of evacuation vehicles \\
$\left[ k\right] = \left(k_1, \dots, k_\ell\right)$ & ordered subtours of vehicle $k$ \\
$K^*$ & set of all vehicle subtours \\
$C_k$ & capacity of the bus $k$\\
$d_i$ & demand of node $i$ \\
$p_i$ & prize of node $i$ \\
$f_i$ & latest pickup time of node $i$ \\
$t_{ij}$ & travel time between node $i$ and $j$ \\
\midrule
\textit{ILP variables} \\
$y_i^{k^*}$ & binary variable to show that vehicle subtour $k^*$ visits node i \\
$x_{ij}^{k^*}$ & binary variable to show that vehicle subtour $k^*$ traverses edge $(i,j)$ \\
$u_i^{k^*}$ & continuous helper variable for vehicle subtour $k^*$ \\
$s_i^{k}$ & continuous helper variable to ensure vehicle $k$ respects time windows \\
$M_T$ & A big M constant (we set $M_T = 10T$) \\
\midrule
\textit{MDP variables} \\
$S^i$ & State of the MDP at step $i$ \\
$\mathcal{V}^i$ & set of evacuated nodes at step $i$ \\
$v^i$ & current location of the vehicle at step $i$ \\
$\lambda^i$ & current load of the vehicle at step $i$ \\
$\tau^i$ & elapsed time of the current vehicle at step $i$ \\
$\kappa^i$ & vehicles to route left at step $i$ \\
$A_s$ & Set of possible actions at state $S$ \\
$a$ & $a \in A$ an action \\
$R$ & reward of terminal state \\
\bottomrule
\end{tabular*}
\end{table}

\subsection{Markov Decision Process for the BEOP}
Complementary to the MILP above, we now provide a Markov decision process (MDP) for the BEOP.
We define the MDP with $(S, A, P, R)$ with a state space $S$, an action space $A$, a probability transition function $P$ and a reward function $R$. We note that our $P$ is generally deterministic \footnote{Although our MDP is typically deterministic, it remains extremely difficult to solve due to the enormous state space induced by the combinatorial nature of the underlying NP-hard problem.} but can become stochastic if we consider stochastic travel times or demands (and corresponding prizes).
Our MDP creates tours for all vehicles sequentially, i.e. we plan the whole trip for the first vehicle before the next vehicle's tour is planned.

At step $i$ we define the state $S^i$ as follows: $$S^i = (\mathcal{V}^i, v^i, \lambda^i, \tau^i, \kappa^i)$$ where $\mathcal{V}^i \subseteq V \setminus \{0\}$ is the evacuation nodes visited already by any vehicle, $v^i \in V$ is the current location of the vehicle, $\lambda^i$ is the current vehicle load, $\tau^i$ is the time the current vehicle has already spent and $\kappa^i$ is the amount of vehicles that have not started their tours yet.

\begin{figure}
    \centering
    \begin{tikzpicture}[
    node distance=3cm and 2cm,
    box/.style={
        rectangle,
        draw=black,
        minimum width=3cm,
        minimum height=1cm,
        align=center
    },
    arrow/.style={->, thick, >=Stealth}
]

\node (A) [box] {Start};
\node (B) [
    box,
    diamond,
    aspect=2,
    right=of A
] {MDP State $S$};
\node (C) [box, right=of B] {End};

\node (D) [box, below=of A, rounded corners] {Evacuation Action: \\ Visit Further \\ Evacuation Node \\ (Case 1)};
\node (E) [box, below=3.5cm of B, rounded corners] {Drop-off Action: \\Return to Safe  \\Location for Drop-off \\ (Case 3)};
\node (F) [box, below=of C, rounded corners] {Terminate Route Action: \\Return to Safe \\Location to End Route \\ (Case 2)};

\draw[arrow] (A) -- node[above]{} (B);

\draw[arrow] (B) to[bend right=15]
node[midway, sloped, above]{time and capacity left}
(D);

\draw[arrow] (D) to[bend right=15]
node[midway, sloped, above]{True}
(B);

\draw[arrow] (B) to[bend right=15]
node[midway, sloped, above]{no time left}
(F);

\draw[arrow] (F) to[bend right=15]
node[midway, sloped, above]{further vehicle left}
(B);

\draw[arrow] (B) to[bend right=15]
node[midway, sloped, above]{only time left}
(E);
\draw[arrow] (B) to[bend right=15]
node[midway, sloped, below]{(no capacity)}
(E);

\draw[arrow] (E) to[bend right=15]
node[midway, sloped, below]{True}
(B);

\draw[arrow] (F) -- 
node[right, align=center, text width=3.5cm, sloped, below]
{no further vehicles left to route}
(C);

\end{tikzpicture}
    \caption{Simplified visualization of the MDP for BEOP showing the different action cases.}
    \label{fig:mdp_actions}
\end{figure}
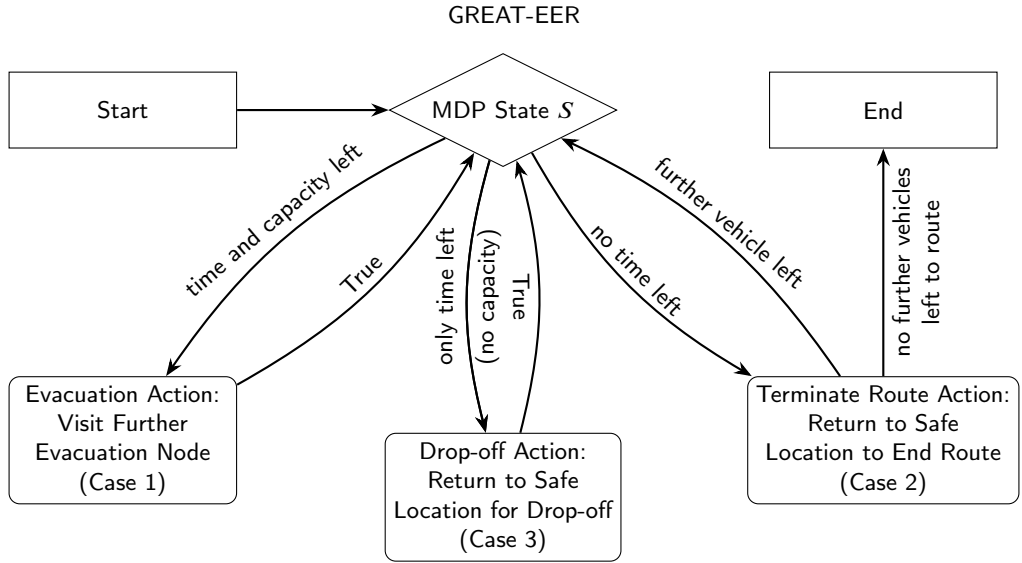

The actions space $A_S$ given a state S is the following: $A_S = \{v \mid v \in V\setminus\{0\}, v \not \in \mathcal{V}^i, d_v + \lambda^i \leq C, t_{v^iv} + t_{v0} + \tau^i \leq T\} \cup \{0\}$, i.e., we can visit any unvisited node whose visitation will not violate the time or capacity constraints or we can return to the depot. A visualization of the idea can be found in Figure \ref{fig:mdp_actions}. Note that Figure \ref{fig:mdp_actions} omits the case where all nodes $v \in V \setminus \{0\}$ have been visited already for the sake of simplicity. In such a case, we would also return to the depot (Case 2) and afterwards terminate.

We start in the initial state $S^0 = (\{\}, 0, 0, 0, |K|\}$. Then, given a state $S^i$ (see above) and an action $a \in A_S$ we have the following state transition:
\begin{enumerate}[label=Case~\arabic*:]
  \item $a \in V \setminus \{ 0 \}$, then $S^{i+1} = (\mathcal{V}^i \cup \{a\}, a, \lambda^i + d_a, \tau^i + t_{v_ia},  \kappa^i)$ 
  \item $a = 0$ and $\not \exists v \in V\setminus \{0\}: v \not \in \mathcal{V}^i, t_{v_ia} + t_{av} + t_{v0} + \tau^i \leq T$ (i.e., there is no other node we can visit in time after executing action $a$), then $S^{i+1} = (\mathcal{V}^i, 0, 0, 0,  \kappa^i - 1)$. This means the plan for the current vehicle is finished and we start the plan for the next vehicle. Note that this is a terminal state if $\kappa^i - 1 = 0$ or $\mathcal{V}^i = V \setminus \{0\}$.
  \item $a = 0$ and $ \exists v \in V\setminus \{0\}: v \not \in \mathcal{V}^i, t_{v_ia} + t_{av} + t_{v0} + \tau^i \leq T$ (i.e., there is another node we can visit in time after executing action $a$), then $S^{i+1} = (\mathcal{V}^i, 0, 0, \tau^i + t_{v_i0},  \kappa^i)$.
\end{enumerate}

Further, we define a reward $R$ for a terminal state $S = (\mathcal{V}, 0, 0, 0, 0)$ as $R = \sum_{n \in \mathcal{V}} p_n$, i.e., the sum of the prizes of all visited nodes.

We note that the above MDP assumes known node demands and travel times. 
In the stochastic case, we restrict the action space using the expected travel time (i.e., if we assume a move will violate the time constraints, we cannot take it). It can happen however, that the expected travel time was to optimistic and the actual value that has been realized by the stochastic distribution violates the maximum travel time. 
In case such a violating happens, we set enter another type of terminal state (representing invalid solutions) and assign a reward of $0$.
In terms of stochastic demands and prizes, we assume that the demand can only be lower than expected (representing people that did not wait for the bus pickup and instead took a car). Then, the vehicle load when transitioning into the updated state does not change (and we do not get a reward for visiting this node in the end either).
We note that in the stochastic case, our MDP works only for a single vehicle, since during real-time operations we have to route all vehicles simultaneously and not subsequently.
Therefore, our MDP can be used for apriori planning of evacuation scenarios with multiple vehicles or real-time evacuation with stochastic components for a single vehicle.

\section{Solution Method}

\subsection{GREAT-EER Framework} \label{sec:framework}
We now introduce the Graph Edge Attention Network for Emergency Evacuation Responses (GREAT-EER) as our framework to generate solutions for the BEOP. A visualization of the framework is shown in Figure \ref{fig:greateer-framework}.
The GREAT-EER framework is an encoder-decoder-based approach. 
This means, that given a suitable input representation (bottom left subfigure in Figure \ref{fig:greateer-framework}) a neural encoding architecture generates hidden feature vectors that represent the edges and nodes in our BEOP instance (second subfigure at the bottom in Figure \ref{fig:greateer-framework}).
Afterwards, a neural decoder architecture selects one node at a time to add to a partial solution tour until a complete solution has been incrementally generated (bottom right subfigure in Figure \ref{fig:greateer-framework}).

In our BEOP setting, the available evacuation time is one of the problem defining properties. 
Consequently, we focus on shortest travel time (compared to travel distance) when deciding on how to route our evacuation vehicles. 
These travel times are an associated edge cost in our routing graph and, in reality, in contrast to Euclidean distance, highly asymmetric due to one way streets and possibly congestion. We refer to Figure \ref{fig:greateer-framework} to observe the highly non-Euclidean distances of the paths that connect the node pairs.
Therefore, to tackle this non-Euclidean setting, we decide to use the graph edge attention network (GREAT) introduced in \cite{lischka2025greatarchitectureedgebasedgraph} as our deep function approximator for the encoder part of our framework since this neural architecture was developed specifically for asymmetric routing problems.
GREAT is graph-neural-network (GNN)-related architecture that operates purely on edge features. 
This is in contrast to ordinary GNNs that usually operate on node features like Euclidean coordinates.
Given an edge-based input graph with associated edge-features, GREAT iteratively updates these edge features by aggregating information over adjacent edge neighborhoods in every network layer.
By this, using an attention like mechanism, the network can compare the different edges and determine which edges represent meaningful connections in the routing graph to obtain high-quality solutions. 
In particular, we compute the edge feature $e_{ij}^k$ for edge $(i,j)$ in the input graph in layer $k$ of the network as follows:

$$
e_{ij}^k = W^k_3 \Bigg(\sum_{\ell \in N(i)}\bigg(\alpha_{i\ell}^kW_1^ke_{i \ell}^{k-1} \big|\big| \beta_{i\ell}^kW_2^ke_{\ell i}^{k-1}\bigg) \bigg|\bigg| \sum_{\ell \in N(j)}\bigg(\alpha_{j\ell}^kW_1^ke_{j \ell}^{k-1} \big|\big| \beta_{j\ell}^kW_2^ke_{\ell j}^{k-1}\bigg)\Bigg)
$$
Where $W_1^k, W_2^k, W_3^k$ are learnable weights of layer $k$, $\alpha^k, \beta^k$ are learnable attention scores (for in and outgoing edges of a node), $||$ denotes vector concatenation and $e^{k-1}_{mn}$ denotes the feature vector of an edge $(m,n)$ of layer $k-1$. The different concatenations summarize information of both in- and outgoing edges of both endpoints of an edge.
GREAT also contains residual layers, layer normalizations and feed-forward layers. We refer the interested reader for these details to the original paper.

We note that BEOP also has node features, namely the demand (which also corresponds to the prize) of a node and an optional latest pickup time. Since GREAT does not process any node features, we transform them into edge features as described in \cite{lischka2025greatarchitectureedgebasedgraph}.
If a node $j$ has a demand $d_j$ and a time window $t_j$, we add this feature to every edge $e_{*j}$ that ends in node $j$. Intuitively, this means that taking an edge $e_{ij}$ in the vehicle route will lead us to node $j$ which means we need to respect the constraints of node $j$ when taking such an edge. Therefore, we can interpret these node features as edge features.
While this is omitted in the bottom left subfigure in Figure \ref{fig:greateer-framework} for the sake of simplicity, it can be thought of as coloring each directed edge in the graph with the color of its end node that show the evacuation point demand in this visualization.

In total, an input BEOP instance that shall be processed by GREAT is transformed into the following processable format (\textit{Transformed Input} in Figure \ref{fig:greateer-framework}):
a complete graph with no node features and a $9$ (or $10$ if there are time windows) dimensional initial feature vector associated with each edge.
The initial features of an edge $(i,j)$ encode: 
\begin{enumerate}
    \item The actual travel time between node $i$ and $j$.
    \item The travel time normalized by the maximum travel time.
    \item The travel time normalized by the maximum travel time and the number of available vehicles.
    \item The travel time from $i$ to $j$ compared to the shortest travel time from any node $k$ to $j$.
    \item The travel time from $i$ to $j$ compared to the average travel time from any node $k$ to $j$.
    \item The travel time from $i$ to $j$ compared to the shortest travel time from $i$ to any node $k$.
    \item The travel time from $i$ to $j$ compared to the average travel time from $i$ to any node $k$.
    \item The demand of node $j$ normalized by vehicle capacity.
    \item The travel time from node $j$ to the safe location.
    \item Optional: The normalized latest pickup time of node $j$ w.r.t. to the maximum travel time.
\end{enumerate}

After the GREAT encoder processed the input graph, we obtain an associated edge embedding for every edge in the graph (\textit{Processed Edges} in Figure \ref{fig:greateer-framework}). 

We then transform these edge embeddings into node embeddings, by aggregating all edge embeddings that share an endpoint to a node.
This is done by an attention-based weighted average such that each node encodes the information of its most promising adjacent edge embeddings.
By this, we reduce the number of embeddings from $\mathcal{O}(n^2)$ edge embeddings to $\mathcal{O}(n)$ node embeddings.
The latent feature vectors of the node embeddings encode the non-Euclidean pairwise distances as captured in the processed edge embeddings.

We then pass the node embeddings to multi-pointer decoder network based on \cite{jin2023pointerformer} (\textit{Solution Decoding} in Figure \ref{fig:greateer-framework}).
Given a query, this pointer iteratively outputs a probability for the next node to visit.
The queries are built to reflect the states in the MDP. In particular, they encode the following information:
\begin{enumerate}
    \item The embedding of the current node the vehicle is at.
    \item The embedding of the safe location.
    \item An embeddings of the whole graph (average of all node embeddings).
    \item An embedding of all nodes visited so far (average of the corresponding node embeddings).
    \item An embedding of the current load of the vehicle.
    \item An embedding of the number of vehicles still available.
    \item An embedding of the remaining time of the current vehicle.
\end{enumerate}

A query is then multiplied by \textit{keys} generated from the node embeddings in a fashion similar to the attention mechanism. 
Different from the attention mechanism, where the scores are then used to update embedding values, the scores are here used as probabilities to visit the next node in our solution tour.
To ensure that our generated solutions are valid (i.e., do not violate any vehicle capacity or time constraints or visit nodes twice), we apply masking operations that set the probabilities of invalid moves to zero.

\subsection{Reinforcement Learning}

To train our GREAT-EER framework for BEOP, we use reinforcement learning.
While some supervised frameworks have shown superior performance in recent works \cite{drakulic2023bqncobisimulationquotientingefficient, drakulic2025goalgeneralistcombinatorialoptimization} when tackling routing problems, BEOP does not have any available strong heuristics or even optimal solution approaches to obtain oracle labels for supervised learning. Using a reinforcement learning-based approach eliminates the need for such labels, effectively allowing us to learn in an unsupervised manner.

As a result, we use the MDP formulation described above in a reinforcement learning setting. 
In particular, we define a reward for the output of the GREAT-EER framework introduced in the last section. 
To maximize the amount of people evacuated, we define the reward to reflect this number. 
We note that the GREAT-EER framework in general only outputs valid solutions due to the masking of invalid operations. However, in the stochastic setting where exact travel times are uncertain, it is possible that a vehicle does not respect the maximum travel time. In such a case we set the reward to zero to guide the model to learn not to make any risky moves. 

We use the POMO framework \cite{kwon2020pomo} to obtain a robust baseline during training and increase the performance of the model during inference.
The framework works by executing a random move in the very first step of the MDP. By this, a diverse set of rollouts is explored. During inference, the best solution achieved by a rollout is then returned as the evacuation route.
During training, each rollout's reward is bench-marked against the average reward of all solutions in order to penalize below average solutions and give a positive emphasize on good solutions.

The REINFORCE algorithm \cite{williams1992simple} is used to compute the gradients of the network based on the rewards, the baseline and the probabilities for the solution tours.
In particular, we sum up the log probabilities of the individual decoding steps of the decoder to obtain probabilities for the whole solution tours.

\section{Experiments} \label{sec:experiments}

\subsection{Environment}

For our experiments, we use the Python package \cite{boeing2017osmnx} to access OpenStreetMap \cite{openstreetmap} to obtain a real world graph of San Francisco. The graph contains approximately $10,000$ nodes representing intersections. Further, the graph contains the travel time for all road segments based on its distance and speed limit. We use Dijkstra's Algorithm \cite{dijkstra2022note} to compute the shortest travel time between any pair of nodes, resulting in a $10,000 \times 10,000$ big matrix.
We split the node set randomly into $70\%$ training nodes, $15\%$ validation nodes and $15\%$ testing nodes.
Further, we associate each node with a sampling probability. Each node has a base frequency of $1$. This frequency gets adjusted using the publicly available Uber dataset \cite{uber_gps_analysis}. This dataset contains roughly $25,000$ taxi trajectories. We extract the pickup and delivery points of these trajectories to obtain GPS locations of points of interest in San Francisco. We use these points as an abstract reflection of population density, aiming to make sampling of nodes more probable if the are often frequented by travelers. We map the GPS locations of such taxi start and end points to the nearest location of a node in our OpenStreetMap graph and increment the nodes frequency count by one for each occurrence.
These frequencies are then used to sample subsets of the training, testing and validation node sets.
We sample graphs of fixed sizes (e.g., 101 nodes) to reflect evacuation scenarios with 100 evacuation points and one depot.
Further, we assign random demands between 1 and 5 to each node which reflect the amount of people to evacuate. 
Moreover, we assign random ``latest pickup times'' (i.e. time windows) to some of the evacuation points, reflecting that some people do not want to wait until the end of the evacuation scenario to be picked up due to individual preferences. We assign time windows to up to $30\%$ of evacuation nodes where each time window is drawn uniformly at random from the range $30\%$ to $80\%$ of the maximum evacuation time.

Apart from the deterministic environment described so far, we also consider a stochastic setting. There, the actual travel times and node demands (i.e. evacuees at an evacuation point) are unknown and only revealed during the operation.
In particular, the travel times are sampled from a normal distribution with a $\sigma$ chosen such that $95\%$ of values are withing a $\pm 10\%$ range of the expected travel times. For the node demands, we randomly set $20\%$ of the demands to $0$ which means that the evacuees did not wait for the bus but instead took a car for the evacuation.

\subsection{Hyperparameters and Training}
The hyperparameters of our neural model are chosen in accordance to the original GREAT architecture in \cite{lischka2025greatarchitectureedgebasedgraph} since those parameters perfomed empirically well and offer a reasonable trade-off between expressiveness (parameter number) and runtime.
This means that we use a GREAT encoder with $5$ layers (each layer consisting of an individual GREAT and feedforward sublayer) with hidden dimension $128$ and $8$ attention heads, resulting in approximately $1.4$ million trainable parameters.
The network is trained using the ADAM optimizer \cite{kingma2014adam} with a learning rate of $0.0001$.
Similarly to \cite{kwon2020pomo} and \cite{lischka2025greatarchitectureedgebasedgraph}, we further scale the distances (in our case travel times) with different factors (in the range $0.5$ to $1.5$) in order for the model to learn a more robust distance distribution. This can lead to better generalization performance of unseen BEOP instances and further be used for instance augmentation which allows us to solve a BEOP instance multiple times during inference and keep the best solution found.

\begin{figure}
  \centering
  \subfloat[Deterministic models]{
    \includegraphics[width=0.45\linewidth]{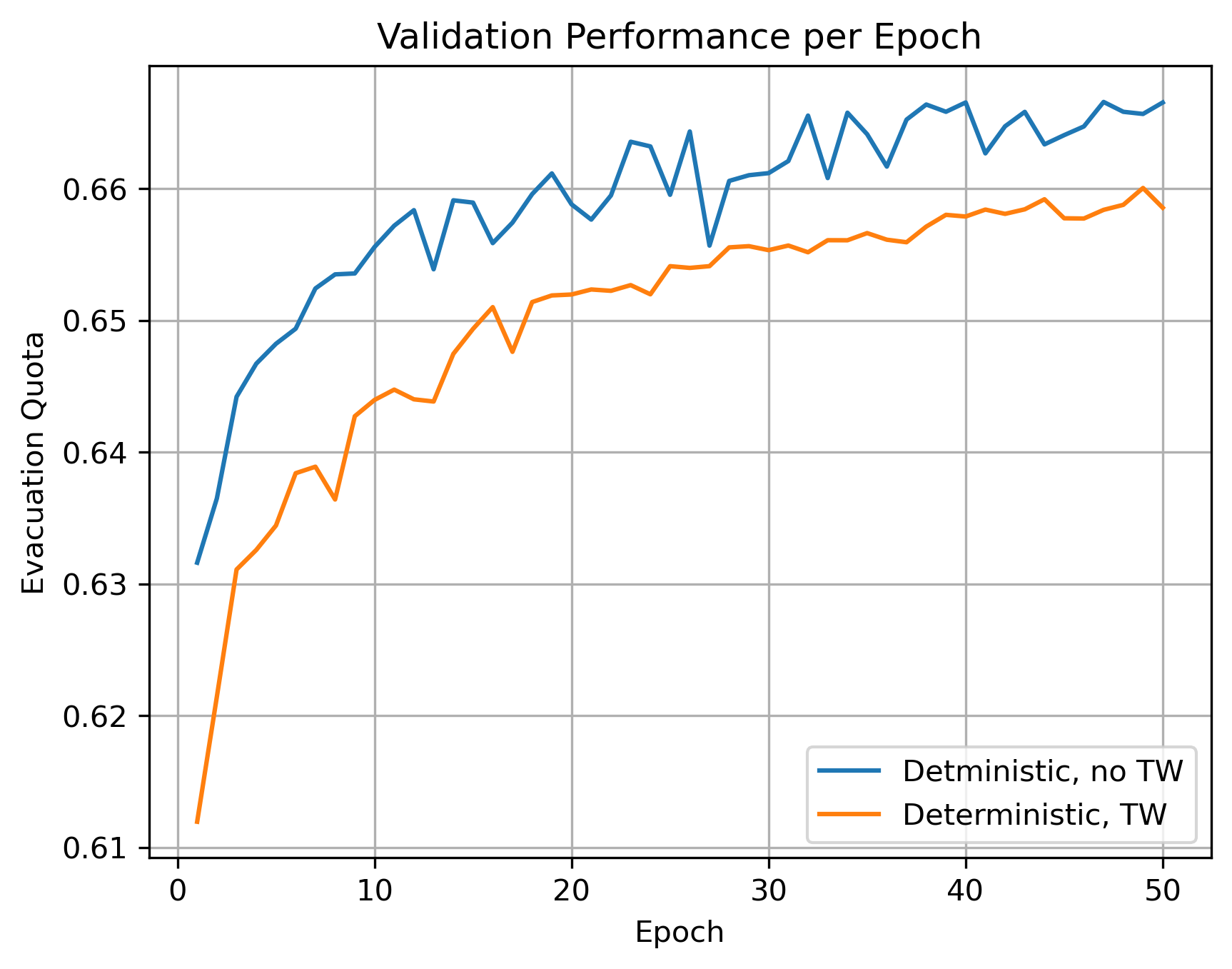}
    \label{fig:img1}
  }
  \hfill
  \subfloat[Stochastic models]{
    \includegraphics[width=0.45\linewidth]{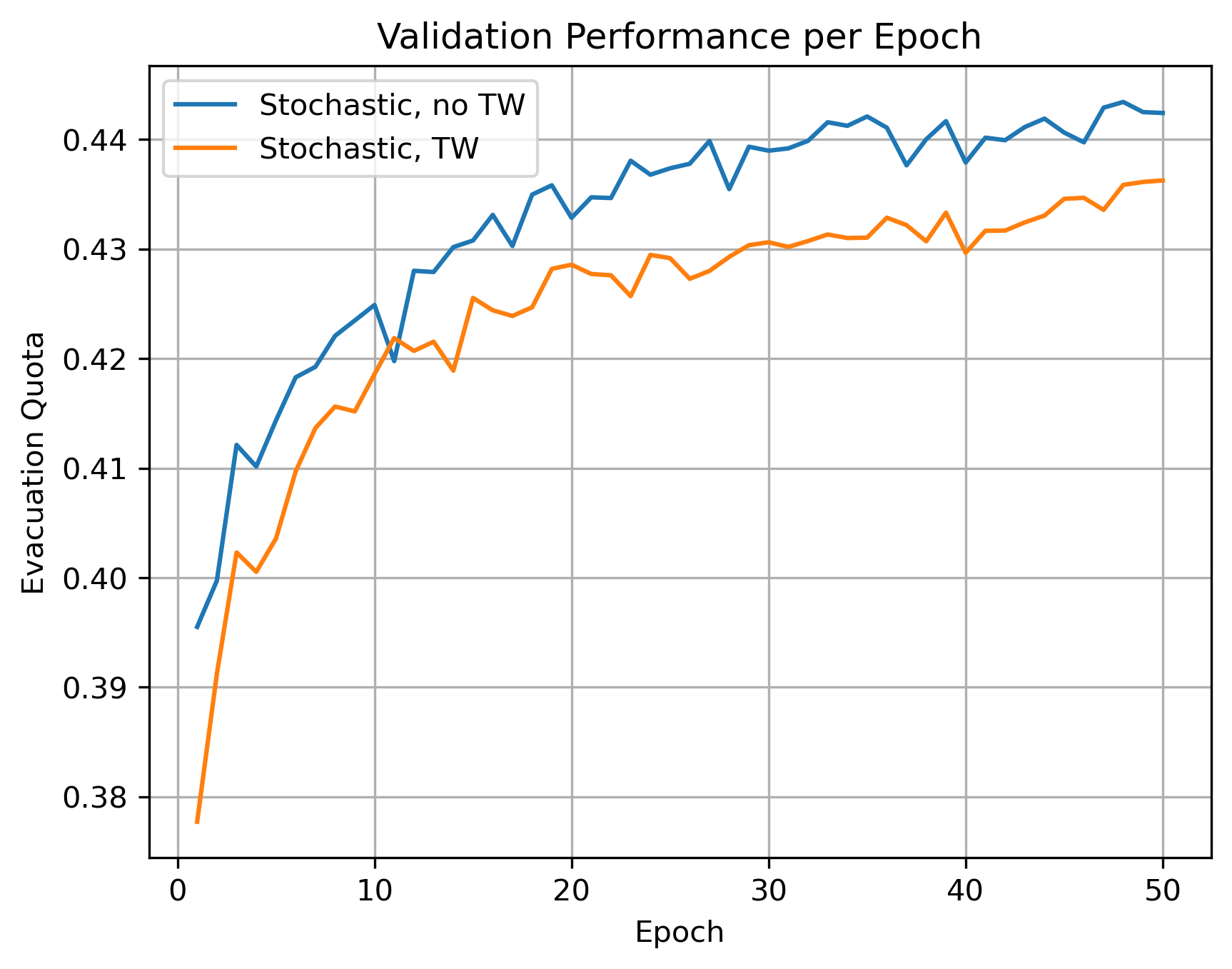}
    \label{fig:img2}
  }
  \caption{Validation performance (evacuation quota) per epoch for GREAT-EER models with 100 evacuation points}
  \label{fig:validation_performance}
\end{figure}

We train four GREAT-EER models (with and without time windows, in the deterministic and stochastic setting) for BEOP instances with 100 evacuation points. 
During training these models, we assume a maximum evacuation times of $0.5, 1$ and $1.5$ hours. In the deterministic setting, we assume that between 1 and 3 buses are available for the evacuation. In the stochastic setting, we consider one vehicle only. Each vehicle has a carrying capacity of 25, 30, 35, 40, 45 or 50 individuals.
The number of vehicles, evacuation time and vehicle capacity are sampled uniformly at random for each BEOP instance during training.
This allows our model to encounter a variety of different problem settings and to learn a robust solution strategy.
Our training dataset contains $25,000$ BEOP instances. The batch size during training is 50 and we train our networks for 50 epochs.
Furthermore, we have a validation dataset of $1,000$ BEOP instances. After each training epoch the model is evaluated on the validation dataset. After the 50 epochs of training, we keep the model that achieved the best performance (i.e. the highest amount of rescued individuals) after any training epoch on the validation dataset. We show the performance of the model (i.e. the average evacuation quota over the $1,000$ validation instances) after each epoch of training in Figure \ref{fig:validation_performance}. The figure shows that all four models can learn successfully, with models operating on BEOP instances without time windows achieving higher performance. We note that the $1,000$ validation are generally a mixture of different BEOP settings (in terms of evacuation time and number of available vehicles). However, since the stochastic setting allows one vehicle only, the curves are shifted down along the y-axis (fewer vehicles naturally lead to lower quotas).
Training is done on NVIDIA A40 GPUs with $48$GB of VRAM, where one epoch of training takes roughly $8.5$ minutes and an additional $1$ minute for validation. As a result the total training time for 50 epochs is $(8.5 + 1) * 50 = 475$ minutes or approximately 8 hours. 

\begin{figure}
    \centering
    \includegraphics[width=0.5\linewidth]{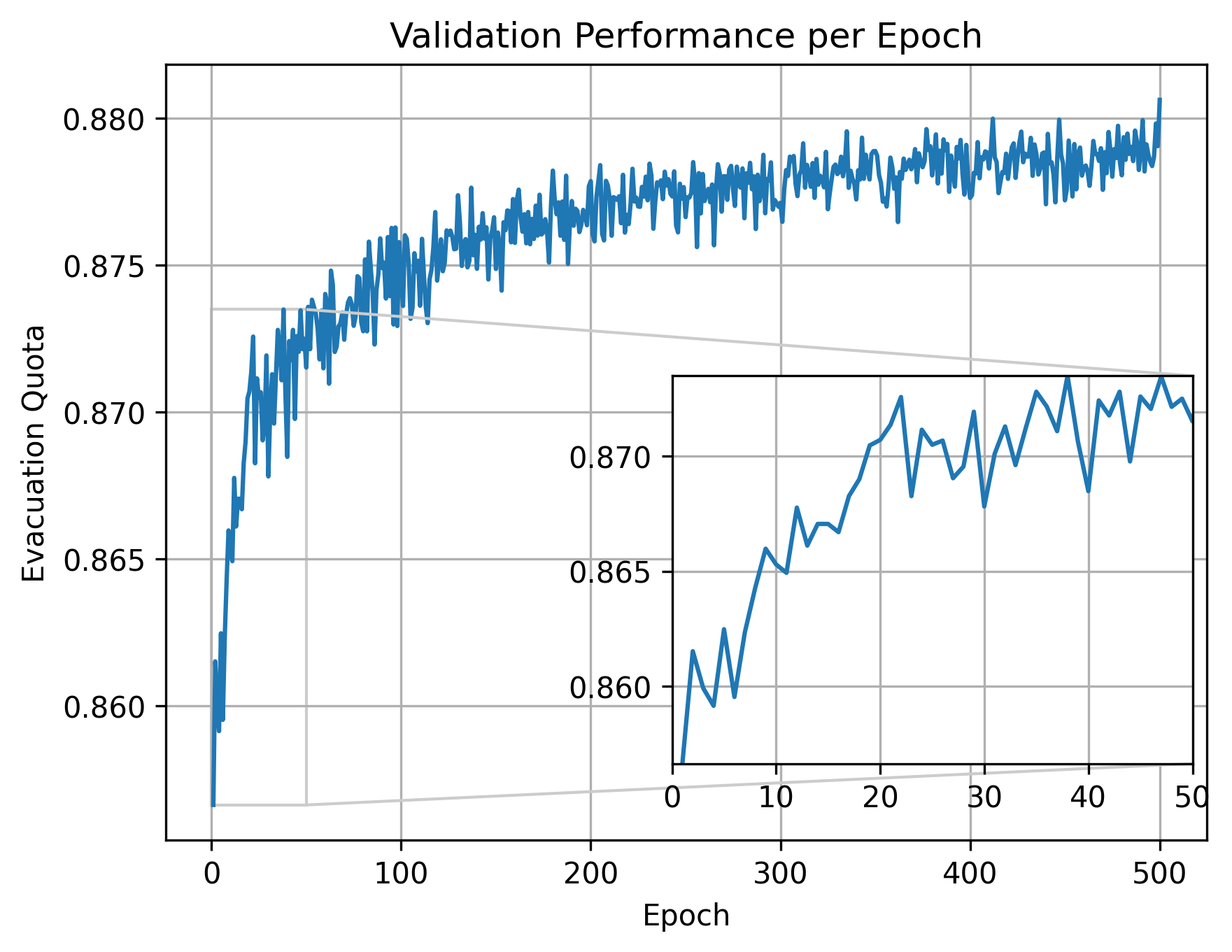}
    \caption{Validation performance (evacuation quota) per epoch for GREAT-EER model with 20 evacuation points}
    \label{fig:validation_performance_small}
\end{figure}

In addition to the training described so far for BEOP instances with 100 evacuation points, we also train a model in a different setting for smaller problem instances of only 20 evacuation nodes. There, we consider instances with at most two busses only, a maximum capacity of 40 and at most 1h of evacuation time to prevent the problems from becoming trivial to solve. This model is trained for 500 epochs on ten datasets of $25,000$ instances (alternating between the datasets in each epoch) which also results in a training time of approximately 8 hours. The validation performance of this small GREAT-EER model is show in  Figure \ref{fig:validation_performance_small}.
When zooming in into the first 50 epochs of the training only, the performance development looks similar to Figure \ref{fig:validation_performance}. However, by keeping training for 8h or 500 epochs we can see that the performance still improves. As a result, we hypothesize that the performance of the larger GREAT-EER models could also be increased further, by prolonging the training time.
Note that we train only one such model in the deterministic setting with time windows, no models without time windows or stochastic models.

The source code of the GREAT-EER framework will be made publicly available after acceptance of the paper.

\subsection{Evaluation}
All evaluations of GREAT-EER models were run on a MacBook Air with M4 Chip. All Gurobi evaluations were run on a computation cluster consisting of Intel Icelake CPUs using 64 cores.

\subsection{Greedy Baseline}

As a baseline, we also introduce a simple greedy heuristic outlined in Algorithm \ref{alg:greedy}. 
Similar to GREAT-EER, the input of the greedy heuristic includes the pair-wise shortest travel times between all nodes in a BEOP instance, as well as evacuation node demand, number of vehicles, vehicle capacity, maximum route length and the number of vehicles. 
The idea of the heuristic is to always greedily visit an unvisited node that has the best tradeoff of ``distance'' (i.e. travel time) and ''prize'' (i.e. number of evacuees at the node) that still respects the maximum travel time, vehicle load and evacuation node time windows.
While this heuristic is naive (and does not provide any bounds w.r.t. optimality), it achieves very fast runtimes (wall-clock time) and serves as a lower bound in terms of performance to compare our GREAT-EER approach to.

\begin{algorithm}[H]
\caption{POSSIBLE\_MOVES (subroutine of greedy heuristic)}
\label{alg:possible_moves}
\begin{algorithmic}[1]
\REQUIRE Current length $L$, return-cost matrix $R$, current node $i$, visited set $V$, current load $q$, demands $d_j$, capacity $C$, time windows $T_j$, normalized distances $D$
\ENSURE Feasible move set $\mathcal{F}$

\STATE $\mathcal{U} \leftarrow$ all nodes $\setminus V$
\STATE $\mathcal{F} \leftarrow \emptyset$

\FOR{each $j \in \mathcal{U}$}
    \STATE $total\_cost \leftarrow L + R_{i,j}$
    \STATE $arrival\_time \leftarrow L + D_{i,j}$
    \IF{$total\_cost \leq 1$ \AND $arrival\_time \leq T_j$ \AND $q + d_j \leq C$}
        \STATE $\mathcal{F} \leftarrow \mathcal{F} \cup \{j\}$
    \ENDIF
\ENDFOR

\RETURN $\mathcal{F}$
\end{algorithmic}
\end{algorithm}

\begin{algorithm}[H]
\caption{Greedy Heuristic}
\label{alg:greedy}
\begin{algorithmic}[1]
\REQUIRE Distance matrix $dist$, prizes/demands $p$, number of vehicles $K$, time windows $T$, capacity $C$, maximum route length $L_{\max}$
\ENSURE Set of tours $\mathcal{T}$

\STATE $D \leftarrow dist / L_{\max}$
\STATE $R_{i,j} \leftarrow D_{i,j} + D_{j,0}$ \COMMENT{travel to $j$ and return to depot}
\STATE $\mathcal{T} \leftarrow \emptyset$

\FOR{$k = 1$ to $K$}
    \STATE $tour \leftarrow [0]$ \COMMENT{start at depot}
    \STATE $L \leftarrow 0$, $q \leftarrow 0$

    \STATE $\mathcal{F} \leftarrow POSSIBLE\_MOVES(L, R, 0, \mathcal{T} \cup tour, q, p, C, T, D)$

    \WHILE{$\mathcal{F} \neq \emptyset$}

        \FOR{each $j \in \mathcal{F}$}
            \STATE $ratio(j) \leftarrow p_j / D_{last(tour),j}$
        \ENDFOR

        \STATE $m \leftarrow \arg\max_{j \in \mathcal{F}} ratio(j)$

        \STATE $L \leftarrow L + D_{last(tour),m}$
        \STATE $q \leftarrow q + p_m$
        \STATE append $m$ to $tour$

        \IF{$m = 0$}
            \STATE $q \leftarrow 0$
        \ENDIF

        \STATE $\mathcal{F} \leftarrow POSSIBLE\_MOVES(L, R, m, \mathcal{T} \cup tour, q, p, C, T, D)$

        \IF{$\mathcal{F} = \emptyset$}
            \STATE $L \leftarrow L + D_{last(tour),0}$
            \STATE $q \leftarrow 0$
            \STATE append $0$ to $tour$
            \STATE $\mathcal{F} \leftarrow POSSIBLE\_MOVES(L, R, 0, \mathcal{T} \cup tour, q, p, C, T, D)$
        \ENDIF

    \ENDWHILE

    \STATE $\mathcal{T} \leftarrow \mathcal{T} \cup tour$

\ENDFOR

\RETURN $\mathcal{T}$
\end{algorithmic}
\end{algorithm}

\subsection{Results}
\begin{figure}
  \centering
  \subfloat[GREAT-EER solution ($94.41\%$ evacuation quota, 1s runtime)]{
    \includegraphics[width=0.45\linewidth]{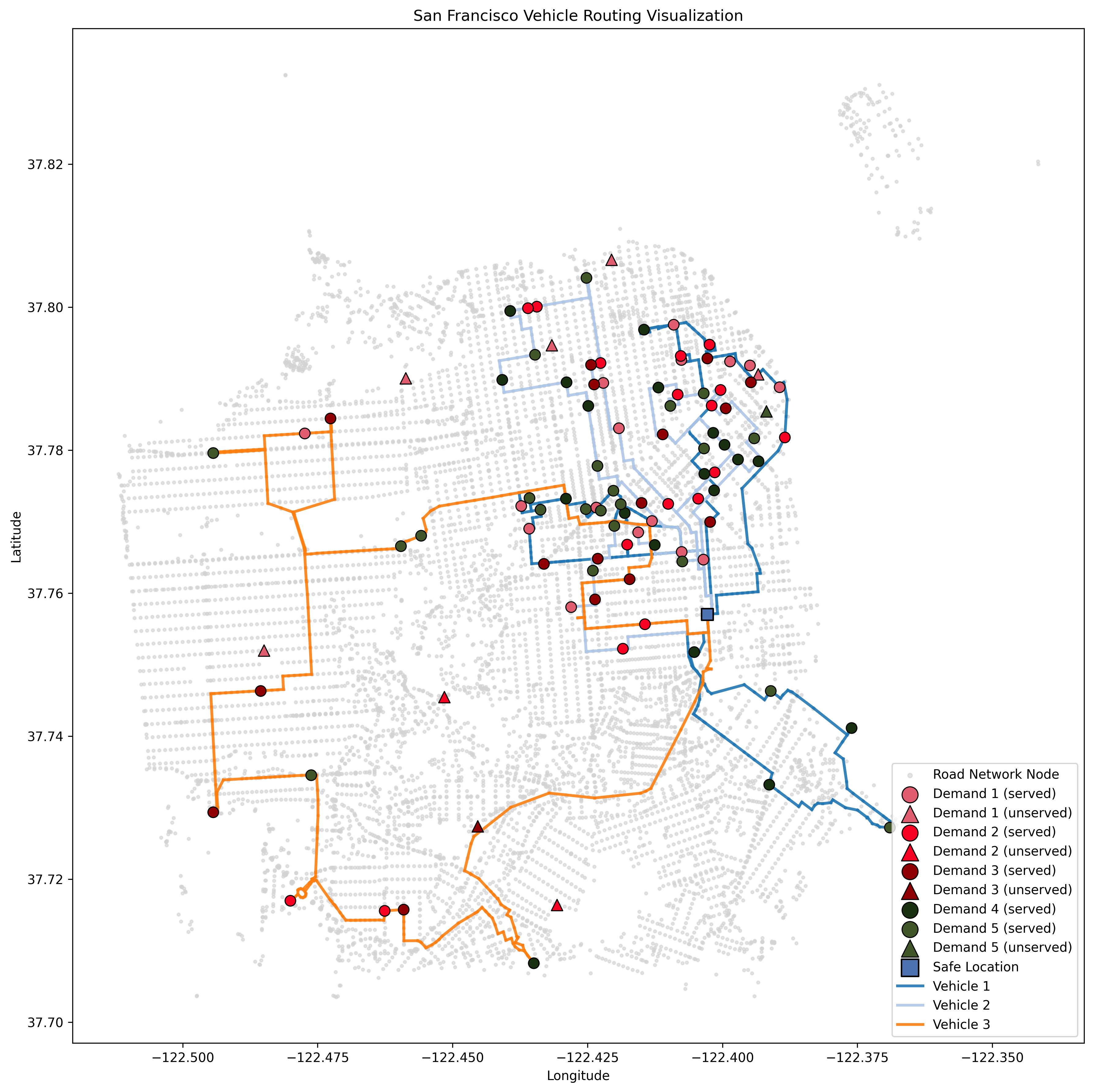}
    \label{fig:img11}
  }
  \hfill
  \subfloat[Greedy heuristic solution ($85.53\%$ evacuation quota, $0.003$s runtime)]{
    \includegraphics[width=0.45\linewidth]{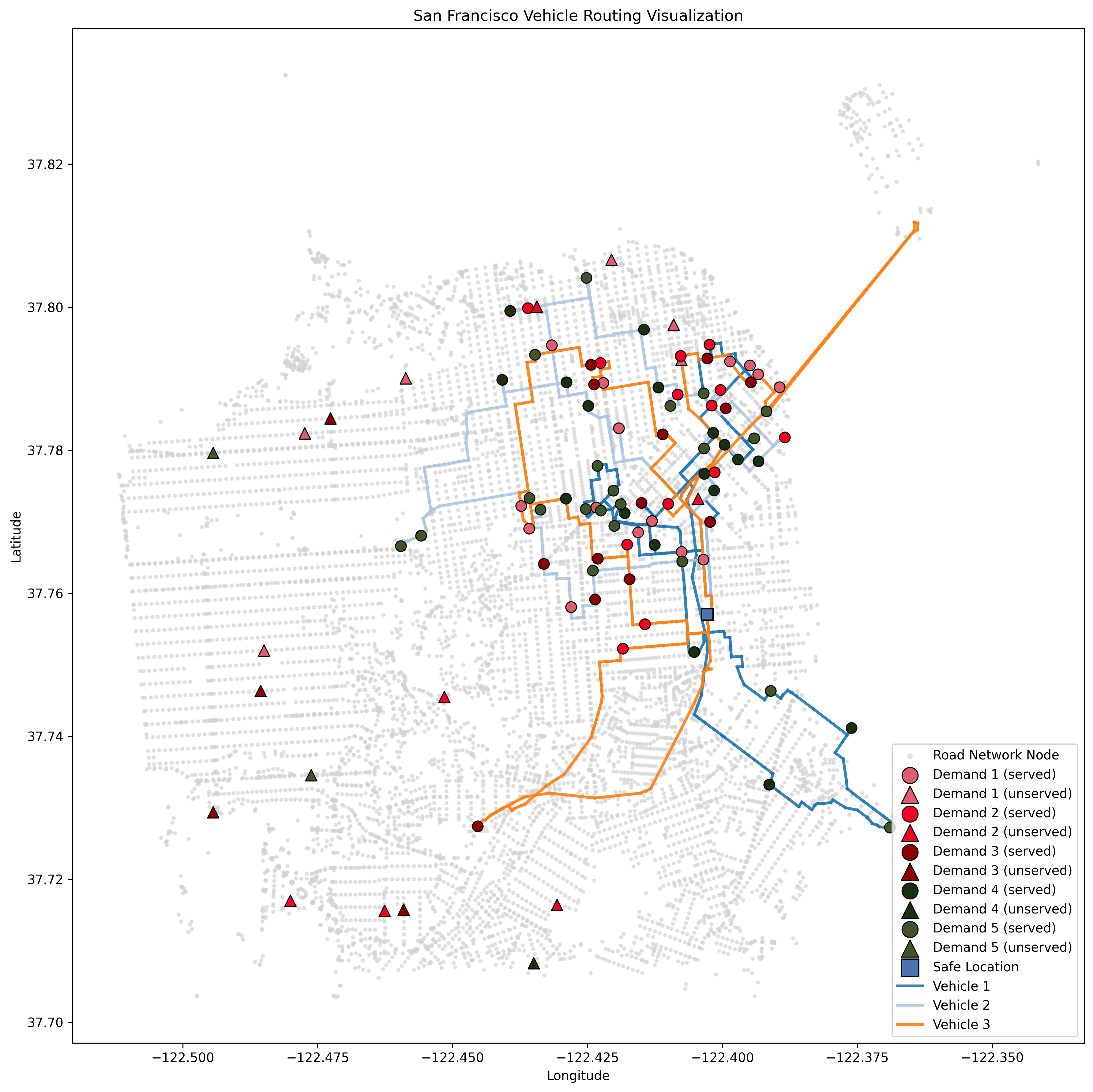}
    \label{fig:img21}
  }
  \hfill
  \subfloat[Gurobi solution with time limit ($88.16 \%$ evacuation quota, 24h runtime)]{
    \includegraphics[width=0.45\linewidth]{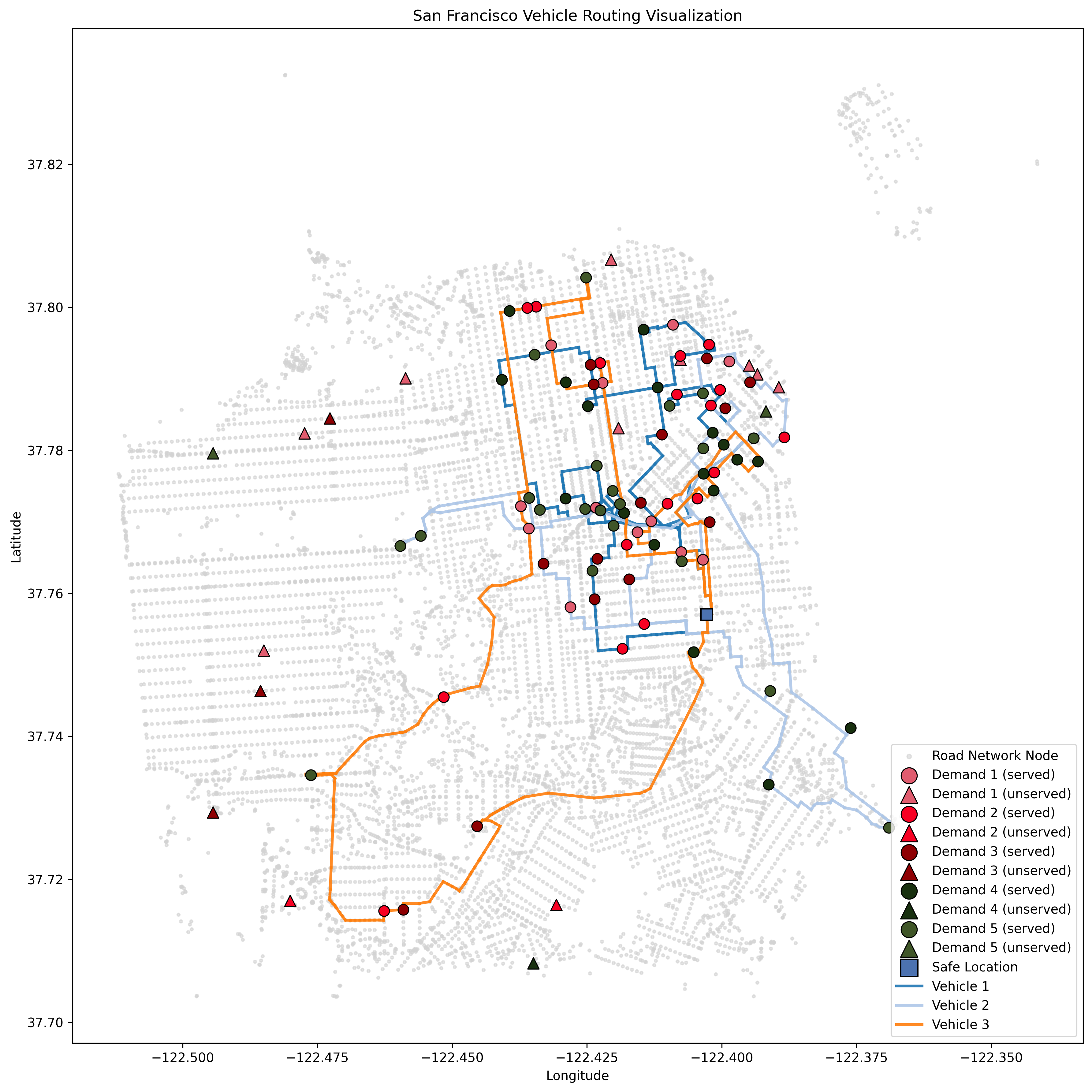}
    \label{fig:img31}
  }
  \hfill
  \subfloat[Gurobi solution with time limit and warm initialization by GREAT-EER ($95.07 \%$ evacuation quota, 10m runtime)]{
    \includegraphics[width=0.45\linewidth]{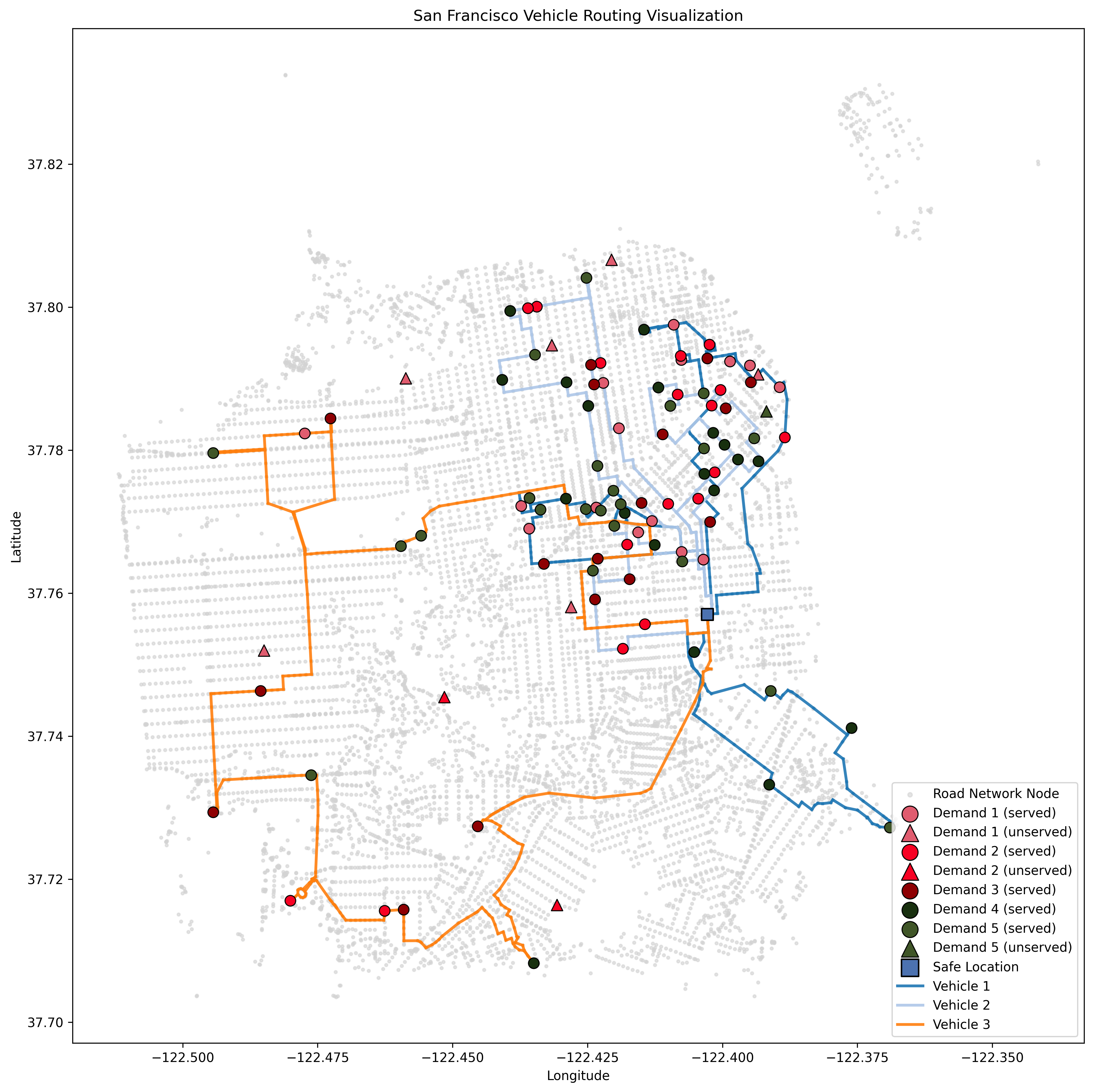}
    \label{fig:img41}
  }
  \caption{Example evacuations generated by different solution methods in San Francisco given 1.5h evacuation time and 3 buses for evacuation with 50 seats each.}
  \label{fig:exampe_evacuations}
\end{figure}

\subsubsection{Deterministic GREAT-EER vs. Greedy vs. MILP model}
As an initial baseline, we consider the small GREAT-EER model that was trained for BEOP instances with only 20 evacuation nodes. 
Due to the relatively small size of these instances, we can compare the performance of the trained GREAT-EER framework not only with the greedy baseline, but also the MILP which is solved using Gurobi. 
We assign a time limit of 30 minutes to each BEOP instance when calling Gurobi. Further, we use the greedy solution as a warm initialization to the problem in this setting. For comparison, we also run Gurobi with a 1 minute time limit with greedy and also the solution provided by GREAT-EER as a warm initialization.
The results of this experiment can be found in Table \ref{tab:great_vs_gurobi} and Table \ref{tab:great_vs_gurobi2}.
To achieve an overview over different BEOP types, we consider instances with 1 or 2 buses (Veh. in the tables), maximum evacuation times (T. in the tables) of $0.5$ and 1h, vehicle capacities of 20 and 40 passengers (Cap. in the tables) and instances with up to 30\% time windows and without time windows (TW in the tables).
For each combination of these parameters, we consider 10 randomly generated BEOP instances in our San Francisco setting. We report the average optimality gap w.r.t. the bound found by the Gurobi model running for 30 minutes, the average evacuation quota and the total runtime (over all 10 instances).

In Table \ref{tab:great_vs_gurobi}, we can see that Gurobi with 30 minutes of runtime consistently achieves the best solutions. We note, however, that the GREAT-EER solution generally achieves almost similar solution quality while only requiring a runtime in the order of fractions of a second ($\sim 0.38$s for solving all ten BEOP instances). We note, that Gurobi sometimes finds optimal solutions before the 30minutes ($1,800$s for 1 instance) are over. While the greedy heuristic requires even less runtime, its performance is lower than the performance achieved by the other solution methods.

\begin{table}[t]
    \centering
    \caption{Comparison of GREAT-EER (G.-EER), Gurobi (30min time limit) and greedy baseline}
    \label{tab:great_vs_gurobi}
    \setlength{\tabcolsep}{6pt}
    \begin{tabular}{cccc|ccc|ccc|ccc}
    \hline
    \multicolumn{4}{c|}{Instance}
    & \multicolumn{3}{c|}{Gap w.r.t. Gurobi Bound(\%)}
    & \multicolumn{3}{c|}{Evac. Quota (\%)}
    & \multicolumn{3}{c}{Runtimes (s)} \\
    Veh & TL & Cap & TW
    & G.-EER & Gurobi & Greedy
    & G.-EER & Gurobi & Greedy
    &  G.-EER & Gurobi & Greedy \\
    \hline
1 & 0.5 & 20 & 0 & 30.19 & \textbf{27.95} & 70.5 & 55.61 & \textbf{56.83} & 44.32 & 0.385 & 10054.9 & 0.003 \\
1 & 0.5 & 20 & 0.3 & 30.36 & \textbf{28.56} & 70.0 & 54.66 & \textbf{56.12} & 43.6 & 0.338 & 9968.1 & 0.002 \\
1 & 0.5 & 40 & 0 & 11.01 & \textbf{8.86} & 29.47 & 68.17 & \textbf{69.61} & 58.6 & 0.351 & 5580.3 & 0.003 \\
1 & 0.5 & 40 & 0.3 & 12.57 & \textbf{9.08} & 31.64 & 66.77 & \textbf{69.05} & 57.25 & 0.351 & 5715.9 & 0.003 \\
1 & 1 & 20 & 0 & 6.57 & \textbf{5.55} & 24.5 & 87.37 & \textbf{88.25} & 75.83 & 0.377 & 11047.5 & 0.004 \\
1 & 1 & 20 & 0.3 & 7.0 & \textbf{5.78} & 26.23 & 87.02 & \textbf{88.06} & 74.84 & 0.373 & 11025.7 & 0.004 \\
1 & 1 & 40 & 0 & 1.62 & \textbf{0.84} & 14.19 & 95.6 & \textbf{96.3} & 85.6 & 0.368 & 7587.2 & 0.004 \\
1 & 1 & 40 & 0.3 & 2.18 & \textbf{1.01} & 14.44 & 94.9 & \textbf{95.96} & 85.2 & 0.373 & 8599.2 & 0.004 \\
2 & 0.5 & 20 & 0 & 15.64 & \textbf{14.59} & 34.11 & 83.95 & \textbf{84.52} & 73.36 & 0.378 & 14560.5 & 0.004 \\
2 & 0.5 & 20 & 0.3 & 15.65 & \textbf{15.43} & 33.84 & 83.78 & \textbf{83.98} & 73.36 & 0.378 & 14549.5 & 0.004 \\
2 & 0.5 & 40 & 0 & 2.81 & \textbf{2.57} & 12.1 & 93.22 & \textbf{93.4} & 85.63 & 0.375 & 12352.7 & 0.004 \\
2 & 0.5 & 40 & 0.3 & 2.76 & \textbf{2.15} & 13.08 & 93.01 & \textbf{93.56} & 84.78 & 0.373 & 12798.6 & 0.004 \\
2 & 1 & 20 & 0 & \textbf{0.0} & \textbf{0.0} & 1.21 & \textbf{100.0} & \textbf{100.0} & 98.85 & 0.389 & 4.6 & 0.005 \\
2 & 1 & 20 & 0.3 & \textbf{0.0} & \textbf{0.0} & 0.68 & \textbf{100.0} & \textbf{100.0} & 99.33 & 0.388 & 6.9 & 0.005 \\
2 & 1 & 40 & 0 & \textbf{0.0} & \textbf{0.0} & 0.21 & \textbf{100.0} & \textbf{100.0} & 99.8 & 0.383 & 3.9 & 0.005 \\
2 & 1 & 40 & 0.3 & \textbf{0.0} & \textbf{0.0} & 0.16 & \textbf{100.0} & \textbf{100.0} & 99.84 & 0.39 & 4.4 & 0.005 \\
\hline
    \end{tabular}
    \end{table}

In comparison, in Table \ref{tab:great_vs_gurobi2} we can see that Gurobi profits considerably when passing the GREAT-EER solution as a warm initialization, outperforming the 30 minute Gurobi baseline in several BEOP settings even though the runtime is limited to 1 minute. Using greedy as the initialization, Gurobi performs similar to GREAT-EER (without Gurobi postprocessing; found in Table \ref{tab:great_vs_gurobi}). In particular, GREAT-EER performs better on instances with 2 vehicles, while Gurobi leads for instances with 1 vehicle only.

As an additional example to show the differences of GREAT-EER, Gurobi and the greedy heuristic, we provide Figure \ref{fig:exampe_evacuations} where an example evacuation scenario of San Francisco with 3 buses of capacity 50 in a $1.5$h time frame is shown. We can see that GREAT-EER achieves a very high evacuation quota of $94.41\%$ in only 1s of runtime. When solving the exact same problem instance with Gurobi (using a time limit of 24h and no warm initialization), only $88.16\%$ of evacuees are picked up. The greedy heuristic achieves only $85.53\%$ evacuation quota but is much faster due to its simplicity. Like on the smaller BEOP instances with only 20 nodes, however, Gurobi can profit significantly when the GREAT-EER solution is passed as a warm initialization. In only 10 minutes of runtime it is then able to improve upon the GREAT-EER solution by visiting an additional node (approximately located at longitude -122.450, latitude 37.73 in the visualizations) and increasing the evacuation quota to $95.07\%$.
We reiterate here that GREAT-EER was only trained for 8h in total and is now applicable to a whole set of BEOP instances while Gurobi needs to be rerun for every BEOP instance and achieves worse performance when running for 24h on a single instance.

\begin{table}[t]
    \centering
    \caption{Comparison of Gurobi (30min time limit), Gurobi (1min time limit) and Gurobi with warm GREAT-EER initialization ``G\&G'' (1min time limit). Gaps w.r.t. the bound found by the Gurobi model with 30 minutes runtime.}
    \label{tab:great_vs_gurobi2}
    \setlength{\tabcolsep}{6pt}
    \begin{tabular}{cccc|ccc|ccc|ccc}
    \hline
    \multicolumn{4}{c|}{Instance}
    & \multicolumn{3}{c|}{Gap w.r.t. Gurobi Bound (\%)}
    & \multicolumn{3}{c}{Evac. Quota (\%)}  & \multicolumn{3}{c}{Runtimes (s)}  \\
    Veh & TL & Cap & TW
    & Gurobi  & Gurobi  & G\&G
     & Gurobi & Gurobi & G\&G & Gurobi & Gurobi & G\&G \\
    & & & & 30min & 1min & 1 min &  30min & 1min & 1min   &  30min & 1min & 1min \\
    \hline
1 & 0.5 & 20 & 0 &  27.95 & 30.23 &  \textbf{26.26} & 56.83 & 55.92 & \textbf{57.12} & 10054.9 & 568.8 & 529.9 \\
1 & 0.5 & 20 & 0.3 &  \textbf{28.56} & 28.98 &  28.79 & \textbf{56.12} & 55.96 & 55.58 & 9968.1 & 558.0 & 561.5 \\
1 & 0.5 & 40 & 0 &  \textbf{8.86} & 9.43 &  \textbf{8.86} & \textbf{69.61} & 69.33 & \textbf{69.61} & 5580.3 & 328.2 & 338.2 \\
1 & 0.5 & 40 & 0.3 &  \textbf{9.08} & \textbf{9.08} &  9.95 & \textbf{69.05} & \textbf{69.05} & 68.34 & 5715.9 & 375.2 & 399.2 \\
1 & 1 & 20 & 0 &  5.55 & 5.55 &  \textbf{5.17} & 88.25 & 88.25 & \textbf{88.54} & 11047.5 & 469.2 & 445.1 \\
1 & 1 & 20 & 0.3 &  \textbf{5.78} & 6.64 &  5.98 & \textbf{88.06} & 87.38 & 87.9 & 11025.7 & 474.4 & 487.8 \\
1 & 1 & 40 & 0 &  \textbf{0.84} & 1.07 &  1.02 & \textbf{96.3} & 96.1 & 96.14 & 7587.2 & 438.1 & 423.4 \\
1 & 1 & 40 & 0.3 &  \textbf{1.01} & 2.56 &  1.19 & \textbf{95.96} & 94.57 & 95.8 & 8599.2 & 482.3 & 419.9 \\
2 & 0.5 & 20 & 0 &  14.59 & 15.94 &  \textbf{14.03} & 84.52 & 83.68 & \textbf{84.94} & 14560.5 & 553.7 & 529.0 \\
2 & 0.5 & 20 & 0.3 &  15.43 & 16.88 &  \textbf{14.93} & 83.98 & 82.95 & \textbf{84.33} & 14549.5 & 559.3 & 525.6 \\
2 & 0.5 & 40 & 0 &  2.57 & 3.67 &  \textbf{2.44} & 93.4 & 92.41 & \textbf{93.54} & 12352.7 & 519.6 & 465.2 \\
2 & 0.5 & 40 & 0.3 &  \textbf{2.15} & 3.07 &  2.2 & \textbf{93.56} & 92.73 & 93.54 & 12798.6 & 536.8 & 484.9 \\
2 & 1 & 20 & 0 &  \textbf{0.0} & \textbf{0.0} &  \textbf{0.0} & \textbf{100.0} & \textbf{100.0} & \textbf{100.0} & 4.6 & 4.6 & 4.8 \\
2 & 1 & 20 & 0.3 &  \textbf{0.0} & \textbf{0.0} &  \textbf{0.0} & \textbf{100.0} & \textbf{100.0} & \textbf{100.0} & 6.9 & 7.0 & 4.9 \\
2 & 1 & 40 & 0 &  \textbf{0.0} & \textbf{0.0} &  \textbf{0.0} & \textbf{100.0} & \textbf{100.0} & \textbf{100.0} & 3.9 & 3.9 & 4.3 \\
2 & 1 & 40 & 0.3 &  \textbf{0.0} & \textbf{0.0} &  \textbf{0.0} & \textbf{100.0} & \textbf{100.0} & \textbf{100.0} & 4.4 & 4.4 & 4.3 \\
\hline
    \end{tabular}
    \end{table}

\subsubsection{Deterministic Evacuation Planning - Quotas}
\begin{figure} %
\centering

\subfloat[1 vehicle - quota vs. capacity]{
  \includegraphics[width=0.38\linewidth]{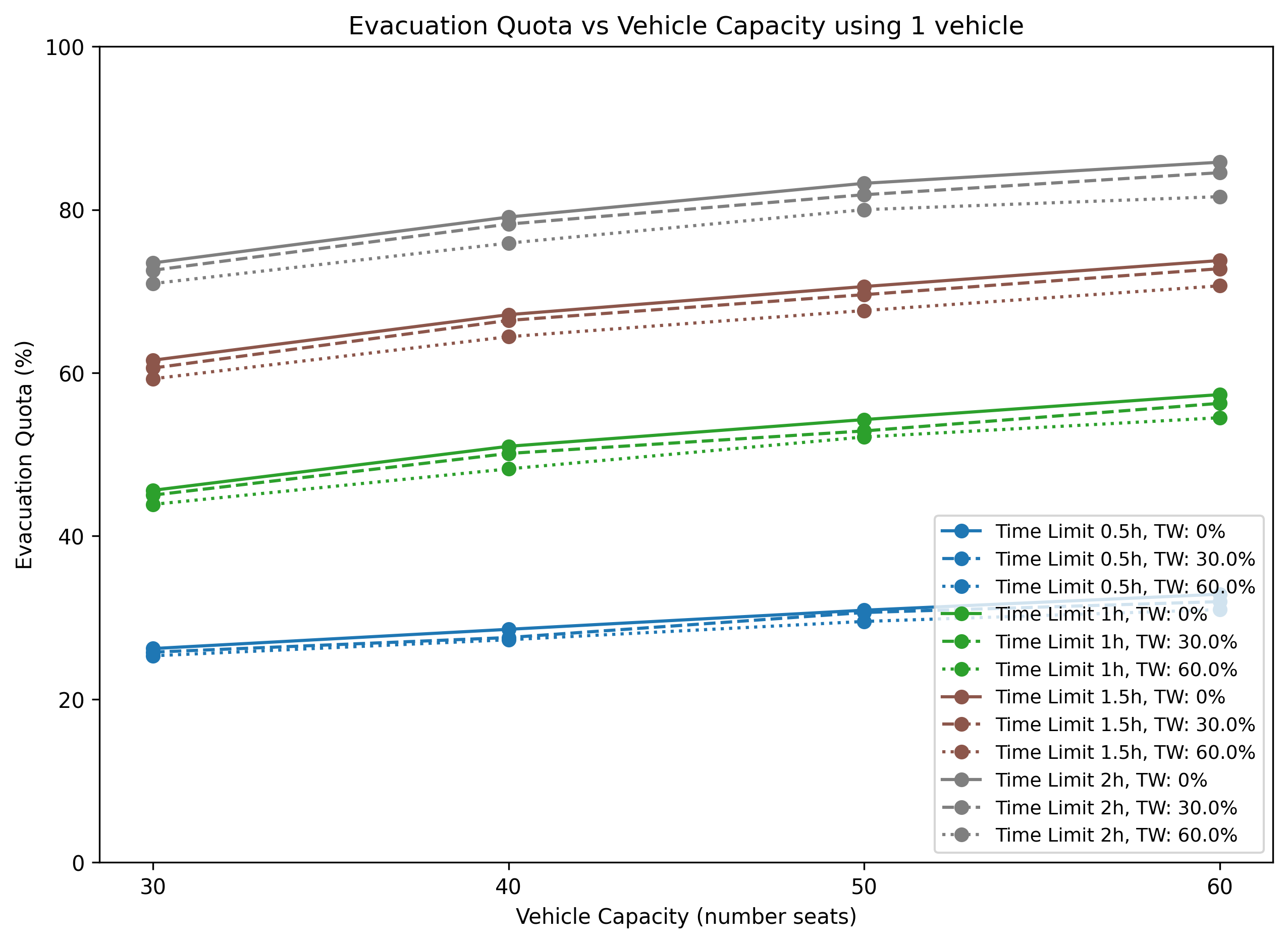}
}\hfill
\subfloat[1 vehicle - quota vs. evacuation time]{
  \includegraphics[width=0.38\linewidth]{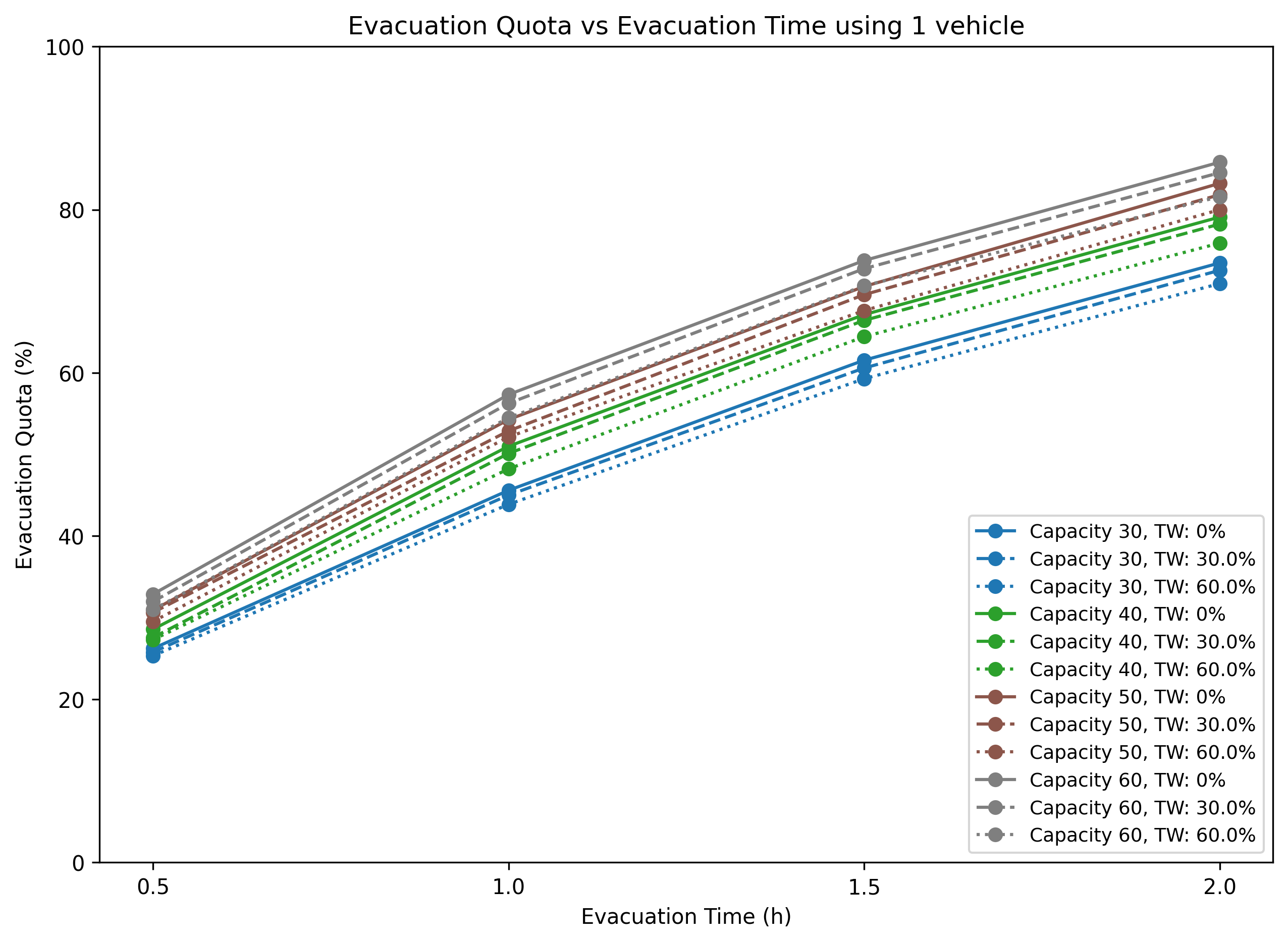}
}\hfill

\subfloat[2 vehicles - quota vs. capacity]{
  \includegraphics[width=0.38\linewidth]{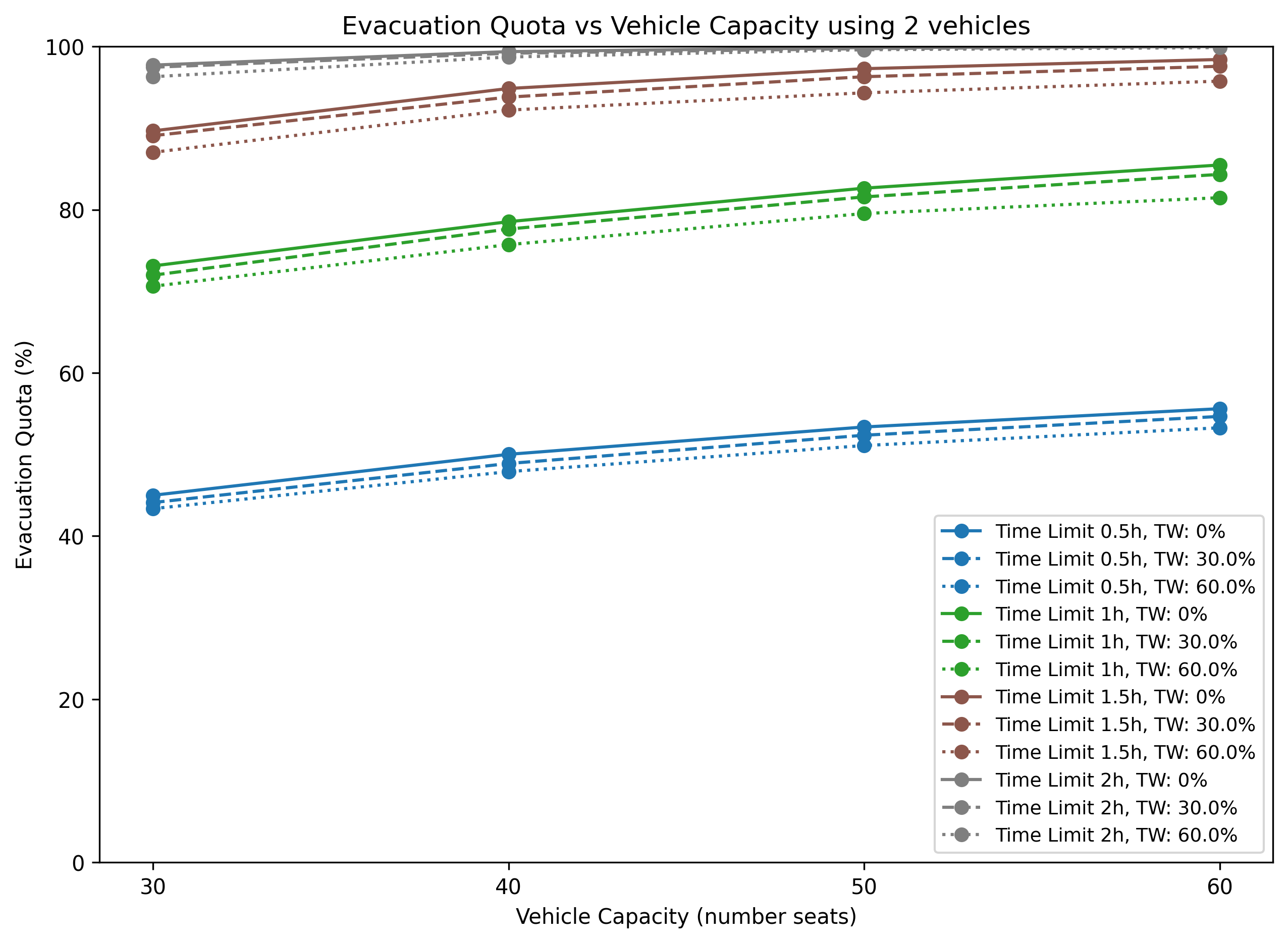}
}\hfill
\subfloat[2 vehicles - quota vs. evacuation time]{
  \includegraphics[width=0.38\linewidth]{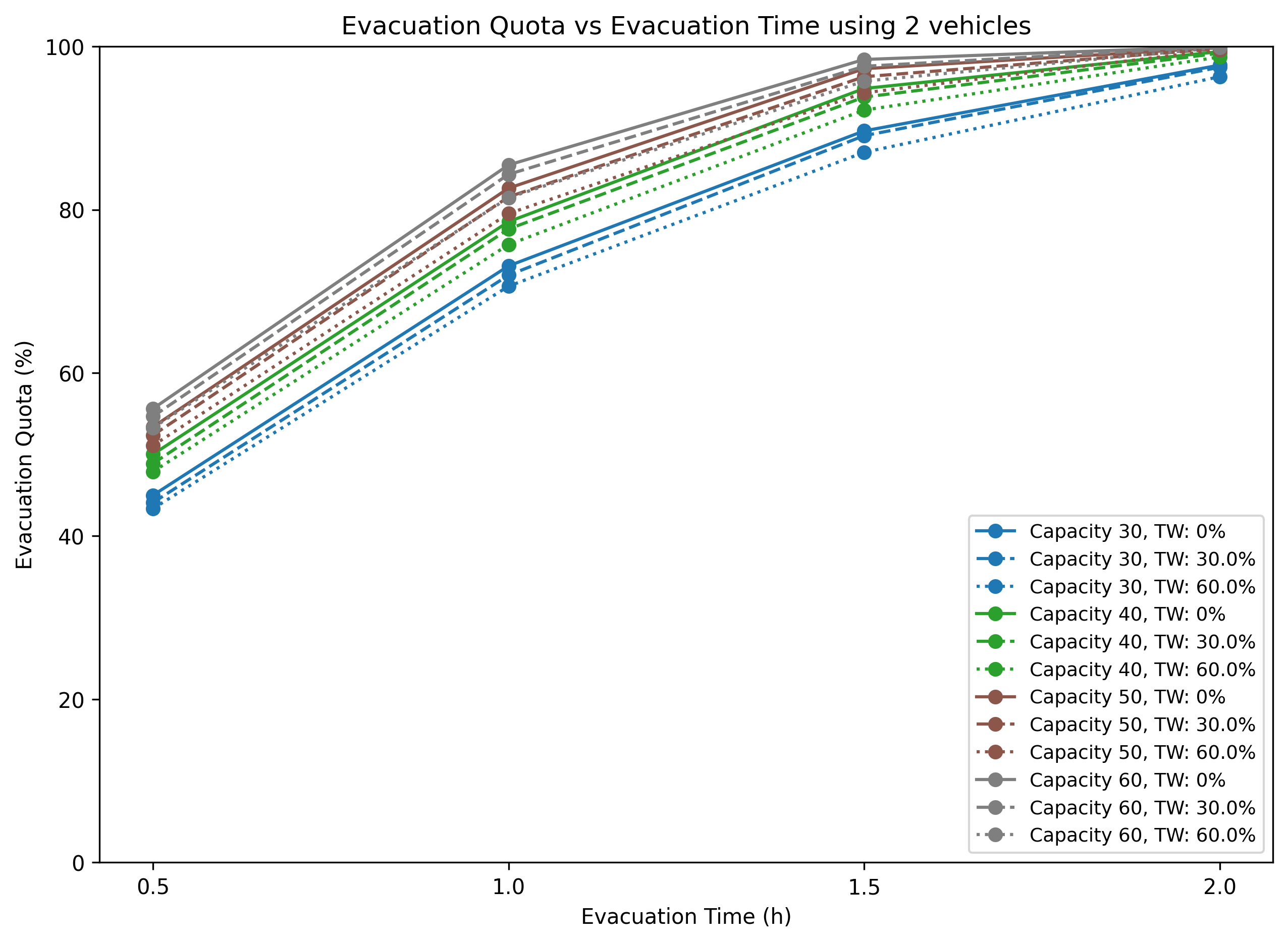}
}\hfill

\subfloat[3 vehicles - quota vs. capacity]{
  \includegraphics[width=0.38\linewidth]{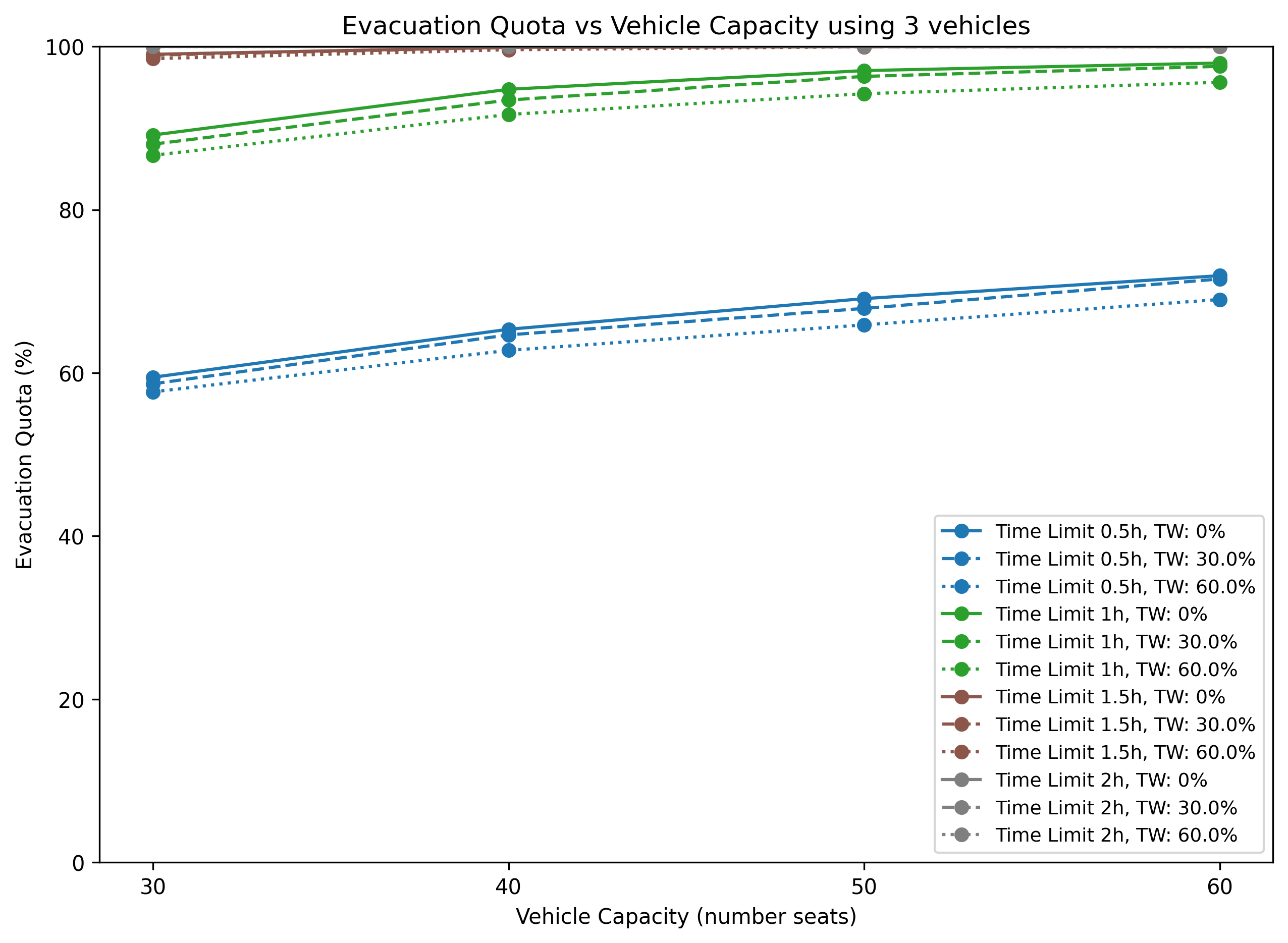}
}\hfill
\subfloat[3 vehicles - quota vs. evacuation time]{
  \includegraphics[width=0.38\linewidth]{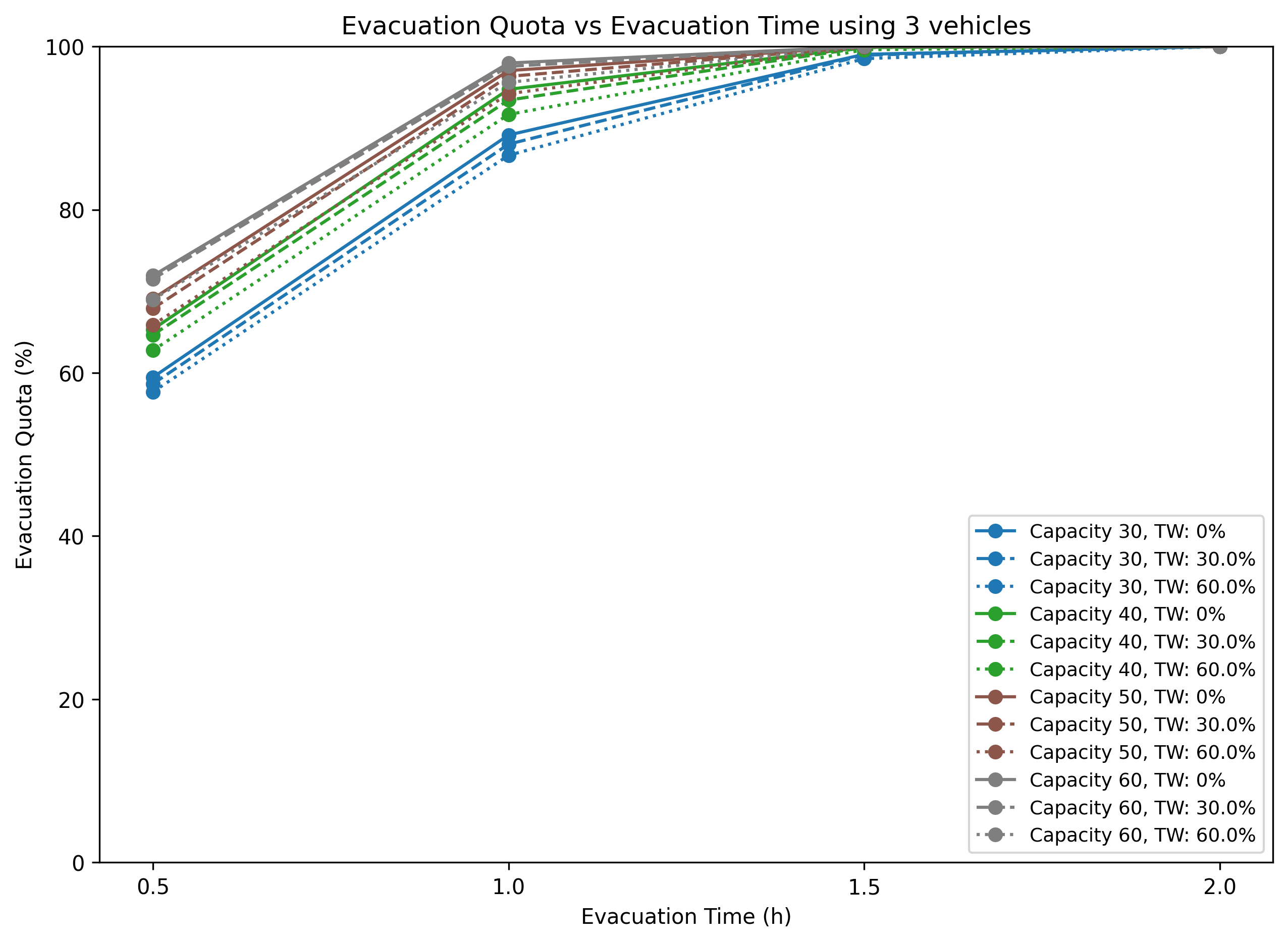}
}\hfill

\subfloat[4 vehicles - quota vs. capacity]{
  \includegraphics[width=0.38\linewidth]{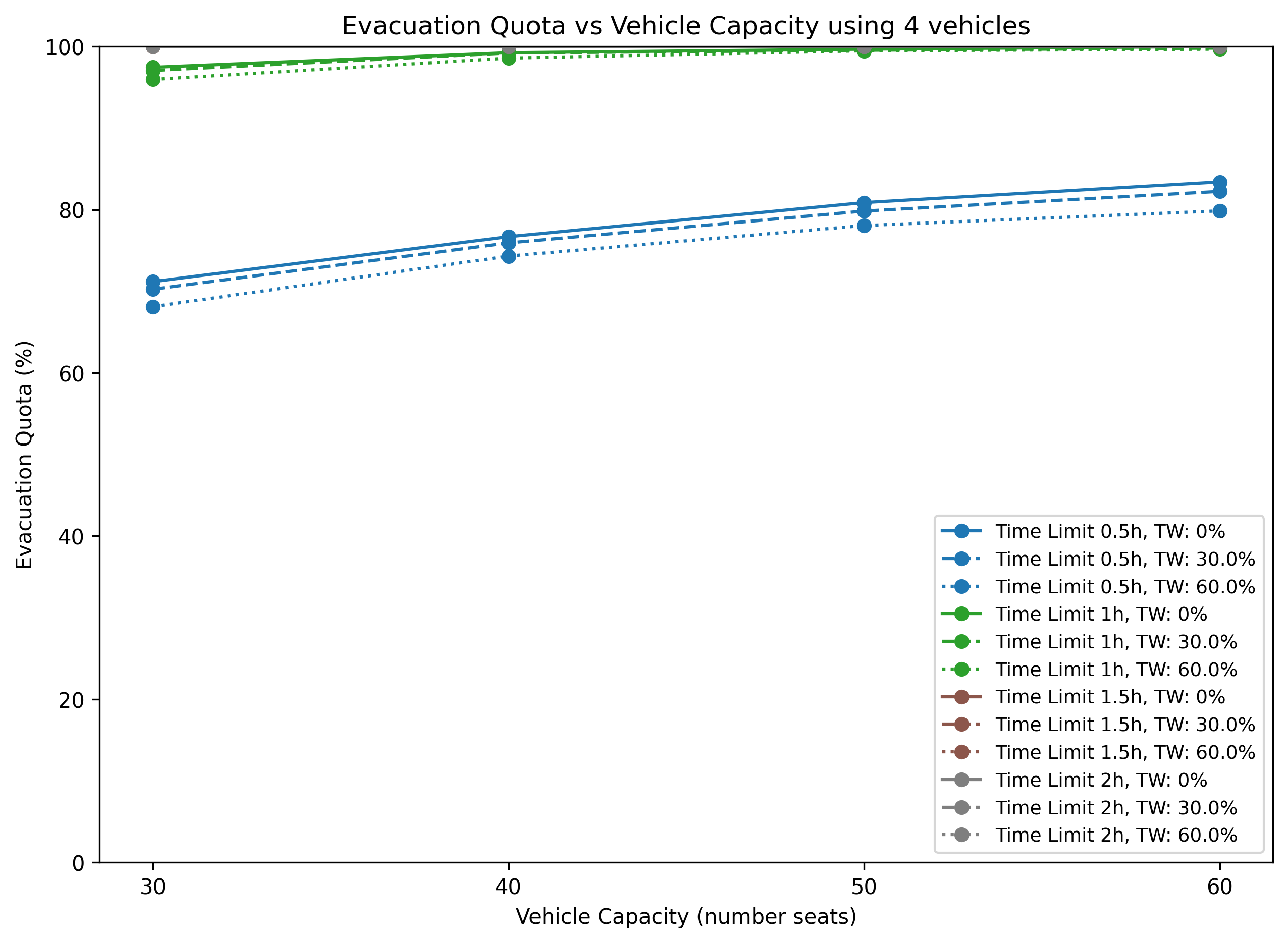}
}\hfill
\subfloat[4 vehicles - quota vs. evacuation time]{
  \includegraphics[width=0.38\linewidth]{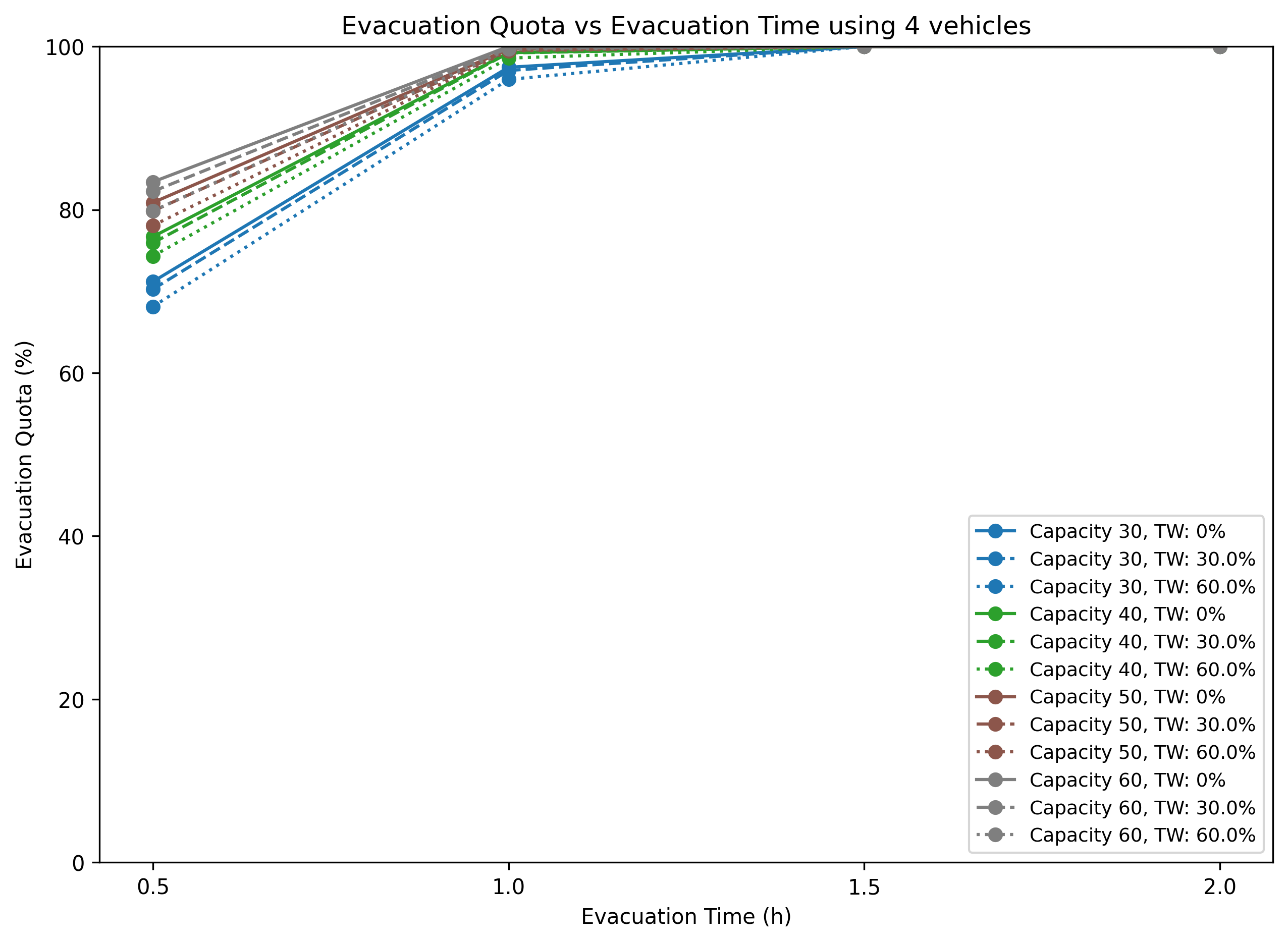}
}

\caption{Evacuation quotas in different BEOP settings (number of vehicles, evacuation time and vehicle capacity)}
\label{fig:8_subfig_grid}
\end{figure}

In this evaluation we use our trained deterministic GREAT-EER model for BEOP instances with 100 evacuation points to derive evacuation quotas given certain settings.
In particular, we investigate how evacuation time, number of available evacuation buses, bus capacity and amount of evacuation points with time windows influence the fraction of people that can be evacuated.
We provide the results in Figure \ref{fig:8_subfig_grid}.
For this evaluation, we use the trained GREAT-EER model for 100 evacuation points.
The trained model is used to compute the evacuation quotas on different BEOP instances, including out-of-distribution (with vehicle capacity greater than 50, more than $1.5$h of evacuation time and more than 3 vehicles). In particular, we consider instances with $0.5, 1, 1.5$ and 2h of evacuation time, $30, 40, 50$ and $60$ vehicle capacity, instances with up to $0.3$ and $0.6$ time window fraction and no time windows. Further, we consider settings with 1 to 4 vehicles. For each combination we consider ten random BEOP instances of the San Francisco setting. 
In addition to the GREAT-EER results, we provide the results by the greedy heuristic as a baseline in Tables \ref{fig:1vehictab} to \ref{fig:4vehictab}. In particular, a table contains two scores for each BEOP instance setting representing the mean evacuation quota achieved by GREAT-EER and greedy, respectively.
We can see, that GREAT-EER outperforms greedy consistently. 
The quota tables and figures allow us to determine how many vehicles are needed in specific settings to achieve certain evacuation quotas. E.g., we can see that we need at least 3 vehicles of 40-50 to achieve a quota of 60\% if the time limit is $0.5$h only.

Overall, using Figure \ref{fig:8_subfig_grid}, we can see that the number of vehicles and the available evacuation time have a much higher influence on the evacuation quota than the amount of evacuation nodes that have time windows and the vehicle capacity. 

\begin{figure} %
\centering

\subfloat[1 vehicle evacuation quota]{
  \includegraphics[width=0.36\linewidth]{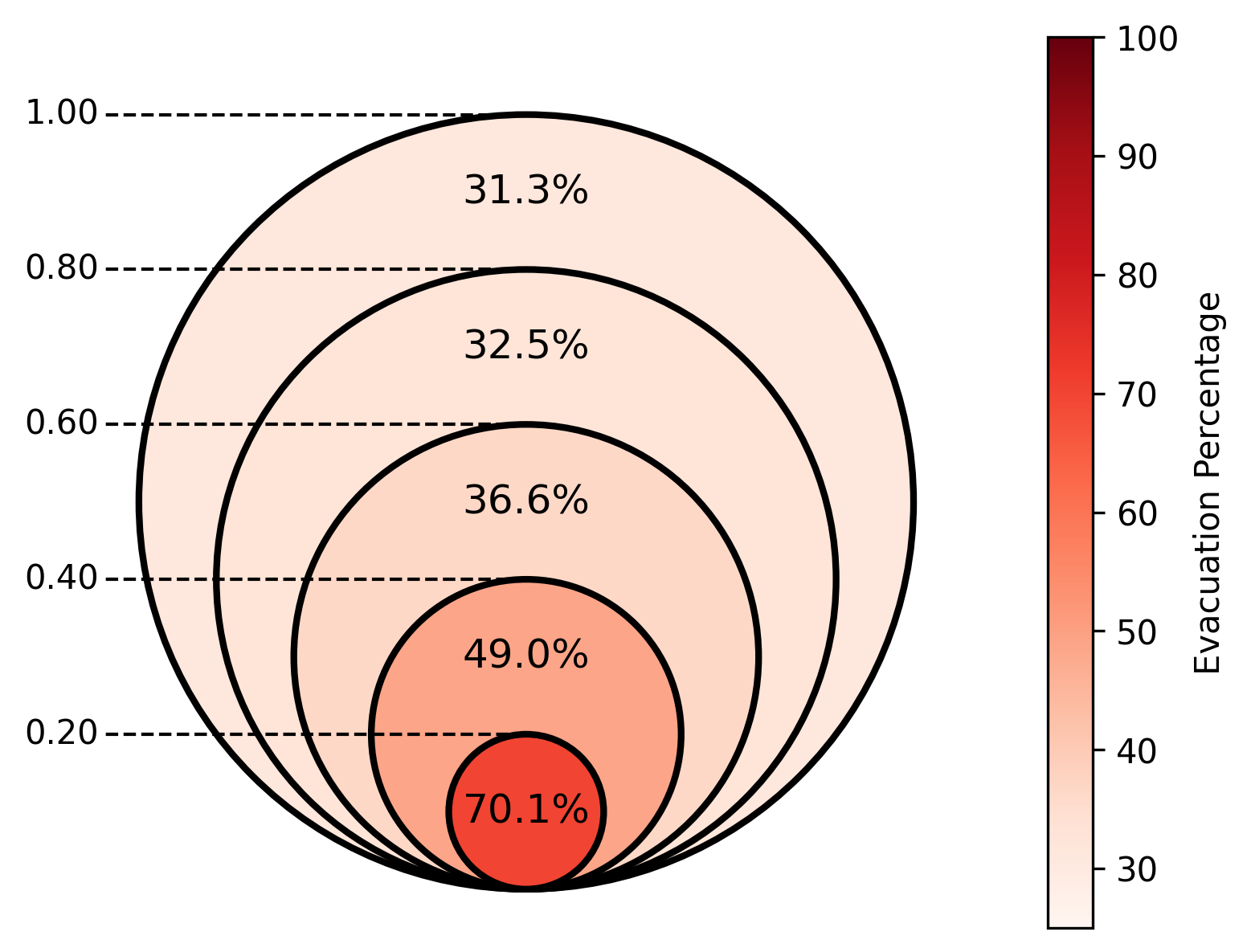}
}\hfill
\subfloat[2 vehicles evacuation quota]{
  \includegraphics[width=0.36\linewidth]{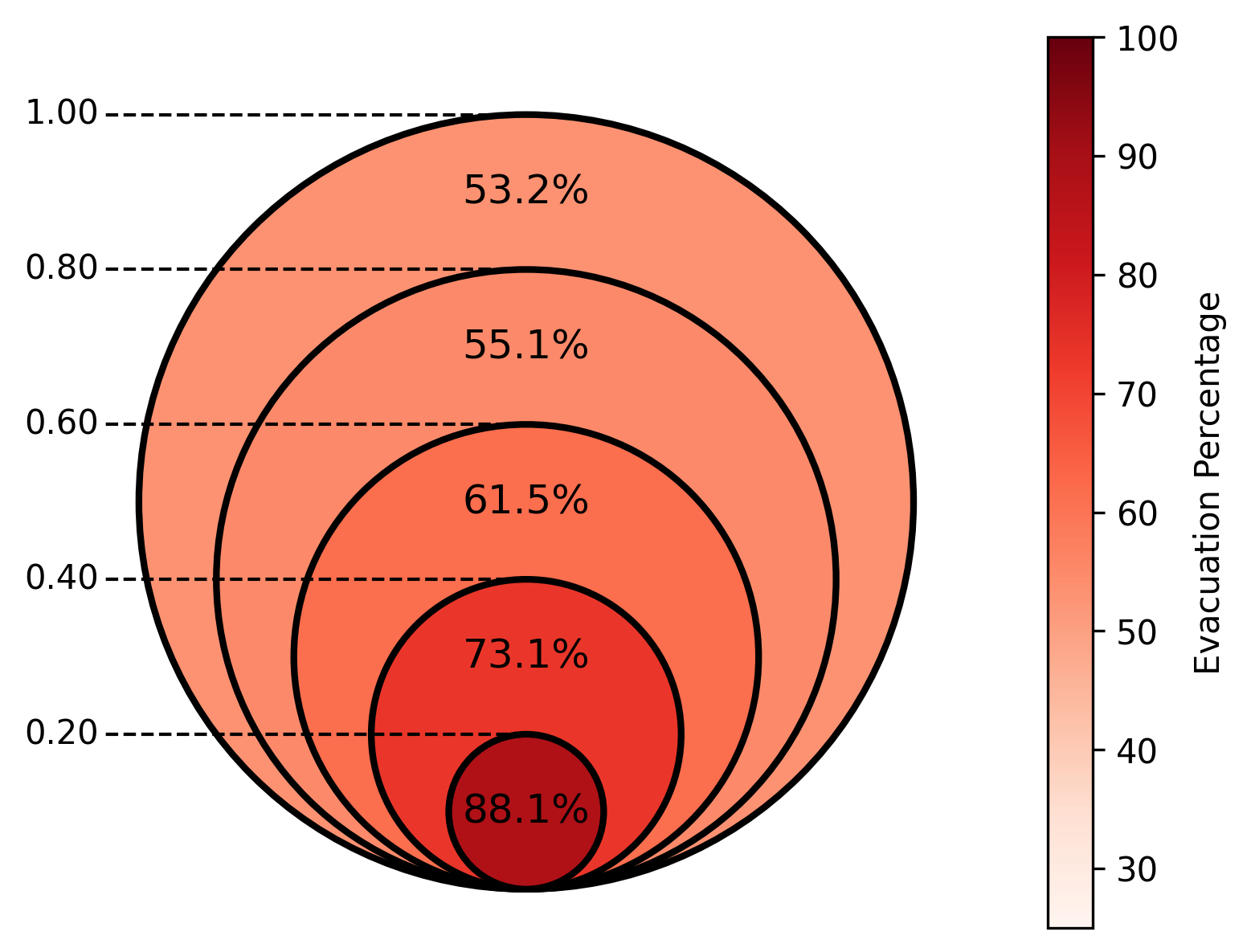}
}\\[1ex]

\subfloat[3 vehicles evacuation quota]{
  \includegraphics[width=0.36\linewidth]{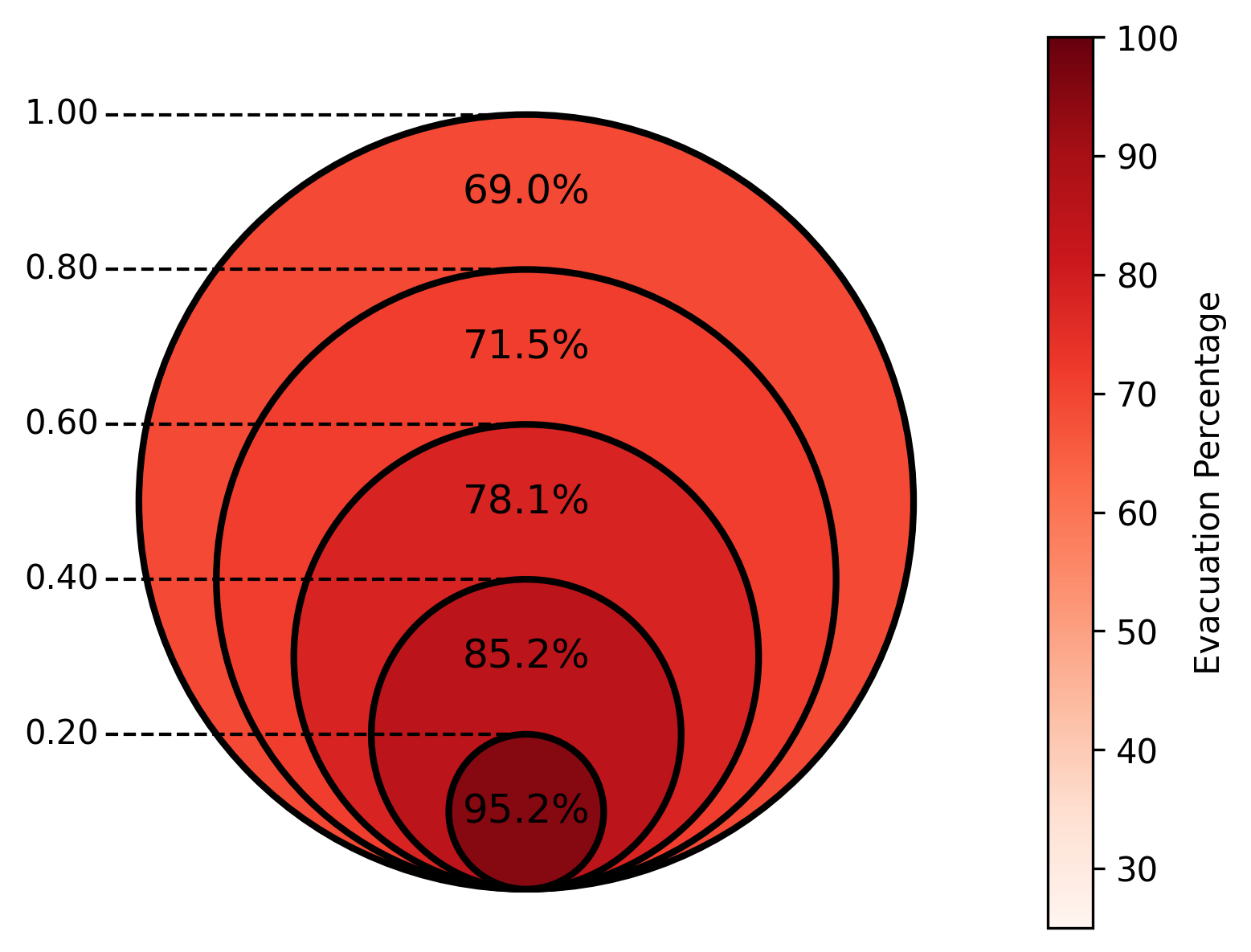}
}\hfill
\subfloat[4 vehicles evacuation quota]{
  \includegraphics[width=0.36\linewidth]{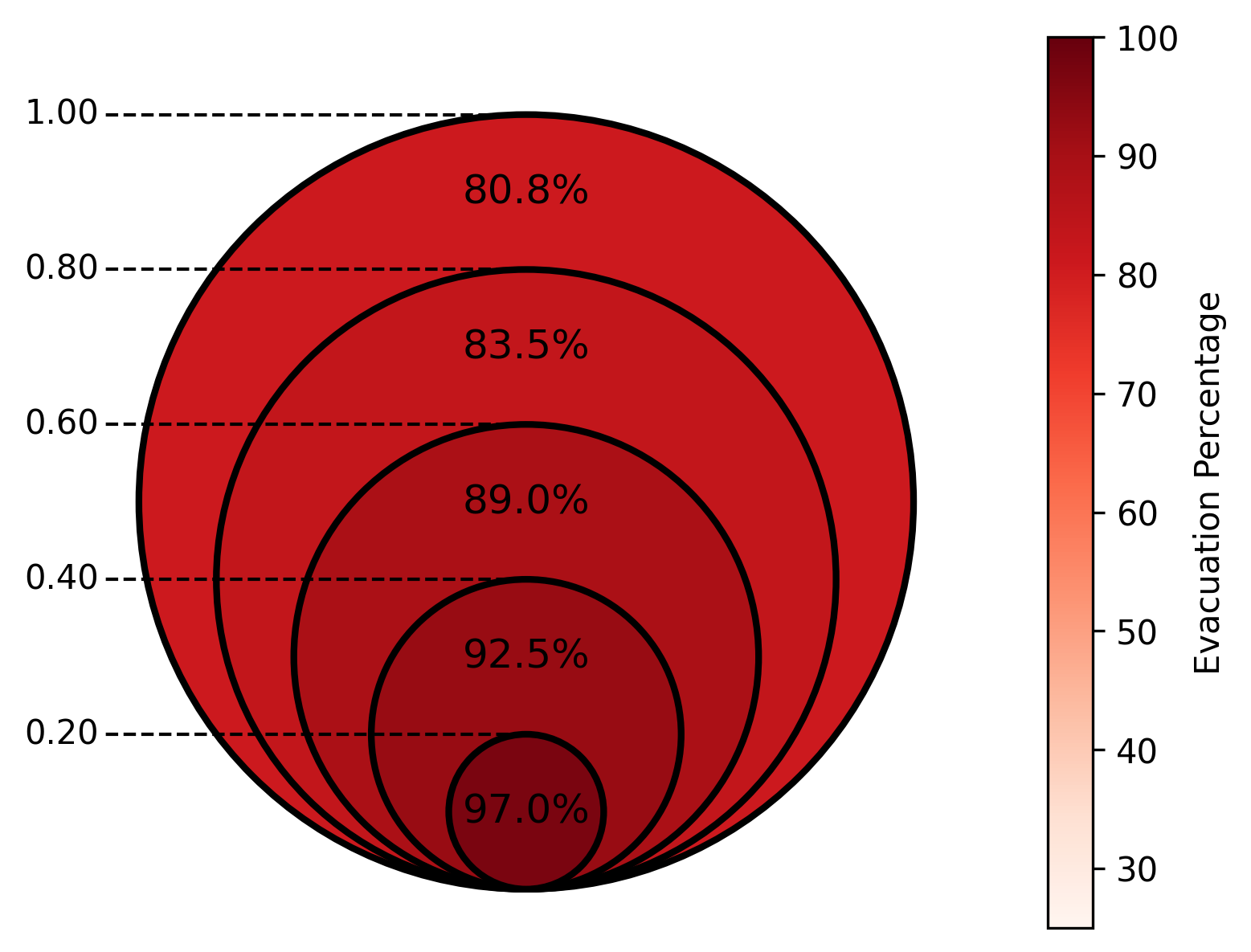}
}\\[1ex]

\caption{Visualization of the quota of people evacuated by bus depending on their distance (fraction of maximum evacuation time) to the safe location.}
\label{fig:8_volumns}
\end{figure}

As an additional insight, we provide a visualization of how the likelihood of evacuees being picked up by bus correlates with their distance to the safe location in Figure \ref{fig:8_volumns}. For this visualization, we consider 100 BEOP instances (per subfigure) with 0.5$h$ of evacuation time and 1-4 buses. A subfigure shows the percentage of people that are evacuated by bus if their distance from the depot is at most a certain fraction ($0.2, 0.4, 0.6, 0.8, 1$) of the total evacuation time. 
We can observe that nodes that are closer to the safe location have a higher evacuation fraction than further away nodes. Moreover, we note that fraction $1$ corresponds to the total evacuation quota.
\begin{table}[ht]
\centering
\caption{Results for $n=1$ vehicles. Entries show GREAT-EER mean evacuation quota and greedy mean evacuation quota.} 
\label{fig:1vehictab}
\begin{tabular}{llcccc}
\toprule
\shortstack{Evacuation\\time} & \shortstack{time window\\fraction} & \shortstack{Capacity \\ 30} & \shortstack{Capacity \\ 40} & \shortstack{Capacity \\ 50} & \shortstack{Capacity \\ 60} \\
\midrule
\multirow{3}{*}{0.5\,h} & 0 & 0.26 / 0.20  & 0.29 / 0.23  & 0.31 / 0.24  & 0.33 / 0.25  \\
 & 0.3 & 0.26 / 0.20  & 0.28 / 0.23  & 0.31 / 0.23  & 0.32 / 0.24  \\
 & 0.6 & 0.25 / 0.19  & 0.27 / 0.21  & 0.30 / 0.22  & 0.31 / 0.23  \\
\midrule
\multirow{3}{*}{1\,h} & 0 & 0.46 / 0.37  & 0.51 / 0.39  & 0.54 / 0.43  & 0.57 / 0.44  \\
 & 0.3 & 0.45 / 0.36  & 0.50 / 0.38  & 0.53 / 0.42  & 0.56 / 0.44  \\
 & 0.6 & 0.44 / 0.36  & 0.48 / 0.38  & 0.52 / 0.42  & 0.55 / 0.43  \\
\midrule
\multirow{3}{*}{1.5\,h} & 0 & 0.62 / 0.50  & 0.67 / 0.55  & 0.71 / 0.57  & 0.74 / 0.59  \\
 & 0.3 & 0.61 / 0.49  & 0.66 / 0.54  & 0.70 / 0.55  & 0.73 / 0.58  \\
 & 0.6 & 0.59 / 0.48  & 0.64 / 0.53  & 0.68 / 0.55  & 0.71 / 0.56  \\
\midrule
\multirow{3}{*}{2\,h} & 0 & 0.73 / 0.62  & 0.79 / 0.68  & 0.83 / 0.72  & 0.86 / 0.74  \\
 & 0.3 & 0.73 / 0.59  & 0.78 / 0.66  & 0.82 / 0.69  & 0.85 / 0.72  \\
 & 0.6 & 0.71 / 0.59  & 0.76 / 0.64  & 0.80 / 0.67  & 0.82 / 0.71  \\
\midrule
\bottomrule
\end{tabular}
\end{table}
\begin{table}[ht]
\centering
\caption{Results for $n=2$ vehicles. Entries show GREAT-EER mean evacuation quota and greedy mean evacuation quota.} 
\begin{tabular}{llcccc}
\toprule
\shortstack{Evacuation\\time} & \shortstack{time window\\fraction} & \shortstack{Capacity \\ 30} & \shortstack{Capacity \\ 40} & \shortstack{Capacity \\ 50} & \shortstack{Capacity \\ 60} \\
\midrule
\multirow{3}{*}{0.5\,h} & 0 & 0.45 / 0.36  & 0.50 / 0.39  & 0.53 / 0.41  & 0.56 / 0.43  \\
 & 0.3 & 0.44 / 0.36  & 0.49 / 0.39  & 0.52 / 0.41  & 0.55 / 0.43  \\
 & 0.6 & 0.43 / 0.34  & 0.48 / 0.37  & 0.51 / 0.40  & 0.53 / 0.42  \\
\midrule
\multirow{3}{*}{1\,h} & 0 & 0.73 / 0.62  & 0.79 / 0.68  & 0.83 / 0.72  & 0.85 / 0.73  \\
 & 0.3 & 0.72 / 0.62  & 0.78 / 0.66  & 0.82 / 0.71  & 0.84 / 0.71  \\
 & 0.6 & 0.71 / 0.60  & 0.76 / 0.64  & 0.80 / 0.69  & 0.81 / 0.69  \\
\midrule
\multirow{3}{*}{1.5\,h} & 0 & 0.90 / 0.79  & 0.95 / 0.85  & 0.97 / 0.88  & 0.98 / 0.91  \\
 & 0.3 & 0.89 / 0.78  & 0.94 / 0.84  & 0.96 / 0.86  & 0.98 / 0.89  \\
 & 0.6 & 0.87 / 0.77  & 0.92 / 0.82  & 0.94 / 0.84  & 0.96 / 0.87  \\
\midrule
\multirow{3}{*}{2\,h} & 0 & 0.98 / 0.92  & 0.99 / 0.96  & 1.00 / 0.98  & 1.00 / 0.99  \\
 & 0.3 & 0.97 / 0.90  & 0.99 / 0.94  & 1.00 / 0.96  & 1.00 / 0.98  \\
 & 0.6 & 0.96 / 0.88  & 0.99 / 0.92  & 1.00 / 0.93  & 1.00 / 0.95  \\
\midrule
\bottomrule
\end{tabular}
\end{table}
\begin{table}[ht]
\centering
\caption{Results for $n=3$ vehicles. Entries show GREAT-EER mean evacuation quota and greedy mean evacuation quota.} 
\begin{tabular}{llcccc}
\toprule
\shortstack{Evacuation\\time} & \shortstack{time window\\fraction} & \shortstack{Capacity \\ 30} & \shortstack{Capacity \\ 40} & \shortstack{Capacity \\ 50} & \shortstack{Capacity \\ 60} \\
\midrule
\multirow{3}{*}{0.5\,h} & 0 & 0.59 / 0.49  & 0.65 / 0.54  & 0.69 / 0.57  & 0.72 / 0.60  \\
 & 0.3 & 0.59 / 0.48  & 0.65 / 0.53  & 0.68 / 0.57  & 0.72 / 0.59  \\
 & 0.6 & 0.58 / 0.47  & 0.63 / 0.51  & 0.66 / 0.56  & 0.69 / 0.58  \\
\midrule
\multirow{3}{*}{1\,h} & 0 & 0.89 / 0.79  & 0.95 / 0.84  & 0.97 / 0.88  & 0.98 / 0.90  \\
 & 0.3 & 0.88 / 0.79  & 0.93 / 0.83  & 0.96 / 0.86  & 0.98 / 0.88  \\
 & 0.6 & 0.87 / 0.77  & 0.92 / 0.80  & 0.94 / 0.85  & 0.96 / 0.86  \\
\midrule
\multirow{3}{*}{1.5\,h} & 0 & 0.99 / 0.95  & 1.00 / 0.98  & 1.00 / 0.99  & 1.00 / 1.00  \\
 & 0.3 & 0.99 / 0.94  & 1.00 / 0.97  & 1.00 / 0.98  & 1.00 / 0.99  \\
 & 0.6 & 0.99 / 0.92  & 1.00 / 0.96  & 1.00 / 0.97  & 1.00 / 0.97  \\
\midrule
\multirow{3}{*}{2\,h} & 0 & 1.00 / 1.00  & 1.00 / 1.00  & 1.00 / 1.00  & 1.00 / 1.00  \\
 & 0.3 & 1.00 / 0.99  & 1.00 / 1.00  & 1.00 / 1.00  & 1.00 / 1.00  \\
 & 0.6 & 1.00 / 0.98  & 1.00 / 1.00  & 1.00 / 1.00  & 1.00 / 1.00  \\
\midrule
\bottomrule
\end{tabular}
\end{table}
\begin{table}[ht]
\centering
\caption{Results for $n=4$ vehicles. Entries show GREAT-EER mean evacuation quota and greedy mean evacuation quota.} 
\label{fig:4vehictab}
\begin{tabular}{llcccc}
\toprule
\shortstack{Evacuation\\time} & \shortstack{time window\\fraction} & \shortstack{Capacity \\ 30} & \shortstack{Capacity \\ 40} & \shortstack{Capacity \\ 50} & \shortstack{Capacity \\ 60} \\
\midrule
\multirow{3}{*}{0.5\,h} & 0 & 0.71 / 0.59  & 0.77 / 0.67  & 0.81 / 0.70  & 0.83 / 0.72  \\
 & 0.3 & 0.70 / 0.59  & 0.76 / 0.65  & 0.80 / 0.69  & 0.82 / 0.70  \\
 & 0.6 & 0.68 / 0.57  & 0.74 / 0.64  & 0.78 / 0.69  & 0.80 / 0.70  \\
\midrule
\multirow{3}{*}{1\,h} & 0 & 0.97 / 0.91  & 0.99 / 0.95  & 1.00 / 0.97  & 1.00 / 0.98  \\
 & 0.3 & 0.97 / 0.89  & 0.99 / 0.94  & 1.00 / 0.97  & 1.00 / 0.97  \\
 & 0.6 & 0.96 / 0.88  & 0.99 / 0.91  & 0.99 / 0.94  & 1.00 / 0.95  \\
\midrule
\multirow{3}{*}{1.5\,h} & 0 & 1.00 / 1.00  & 1.00 / 1.00  & 1.00 / 1.00  & 1.00 / 1.00  \\
 & 0.3 & 1.00 / 0.99  & 1.00 / 1.00  & 1.00 / 1.00  & 1.00 / 1.00  \\
 & 0.6 & 1.00 / 0.99  & 1.00 / 1.00  & 1.00 / 1.00  & 1.00 / 1.00  \\
\midrule
\multirow{3}{*}{2\,h} & 0 & 1.00 / 1.00  & 1.00 / 1.00  & 1.00 / 1.00  & 1.00 / 1.00  \\
 & 0.3 & 1.00 / 1.00  & 1.00 / 1.00  & 1.00 / 1.00  & 1.00 / 1.00  \\
 & 0.6 & 1.00 / 1.00  & 1.00 / 1.00  & 1.00 / 1.00  & 1.00 / 1.00  \\
\midrule
\bottomrule
\end{tabular}
\end{table}

\subsubsection{Out-of-Distribution Results}

In the following, we explore how GREAT-EER performs in distributions different from the training data.
In particular, we explore how the model behaves when parts of the city become inaccessible (e.g. due to wildfires or inundations).
In this setting there are no evacuation nodes in the affected part of the city. Furthermore, it is also not possible to take any roads that pass through this hazard zone. 
Therefore, the distances between nodes are different from the usual case since the fastest route might be blocked. 
The hazard zones we consider have a size such that the affected area it covers would usually contain around 20 evacuation points. 
Instances in this setting are sampled by choosing $121$ nodes in our described San Francisco setting. We chose one node randomly as the safe location and a further random node as the hazard zone epicenter. We then determine the radius around the epicenter such that 20 nodes out of the 121 are affected. 
Afterwards, we remove all roads and road network nodes that are within this radius. We keep the 100 unaffected nodes as our evacuation points.
If the safe location is within the hazard zone or the graph becomes disconnected, we resample.

\begin{figure}
  \centering
  \subfloat[GREAT-EER solution ($95.41\%$ evacuation quota)]{
    \includegraphics[width=0.45\linewidth]{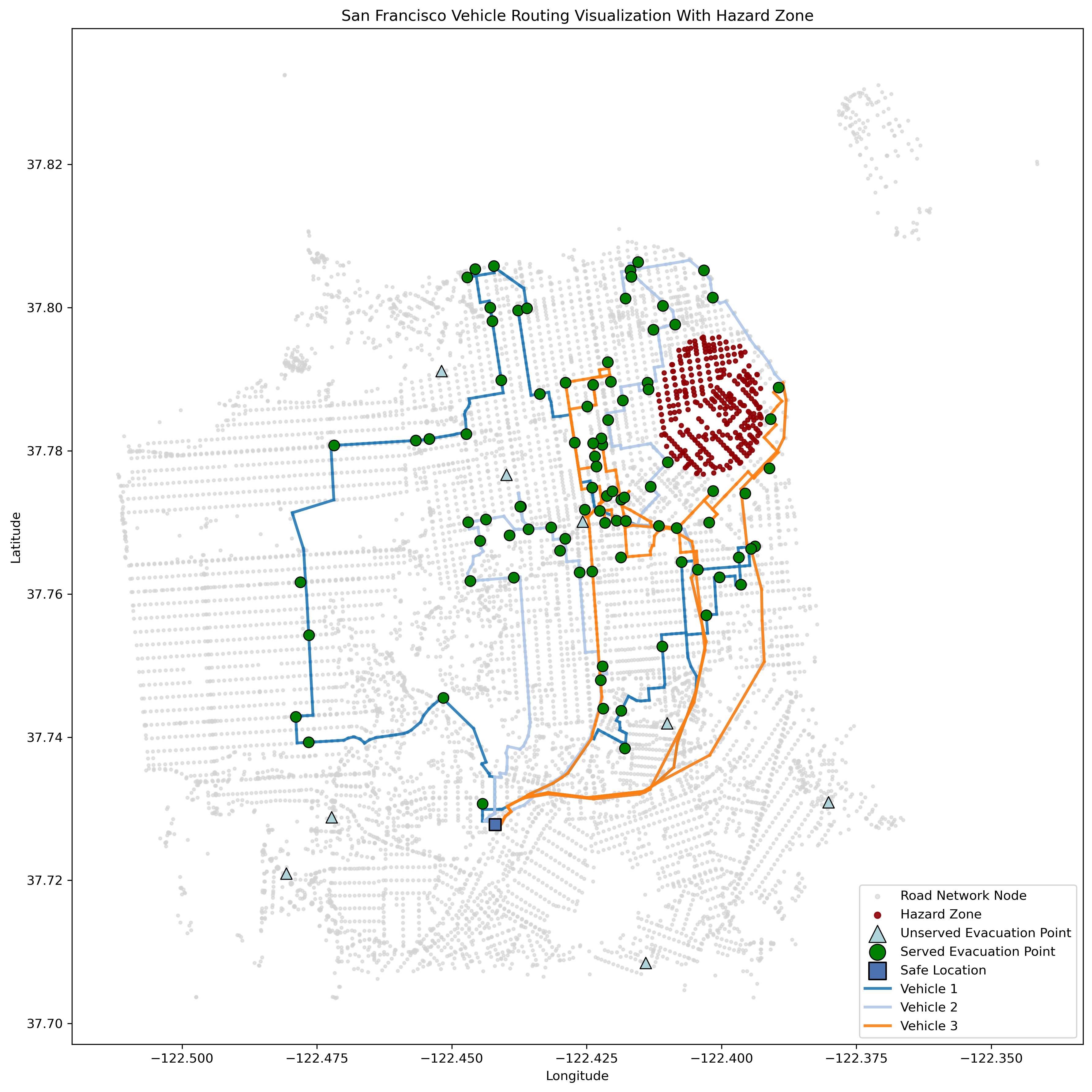}
    \label{fig:img112}
  }
  \hfill
  \subfloat[Greedy solution ($83.39\%$ evacuation quota)]{
    \includegraphics[width=0.45\linewidth]{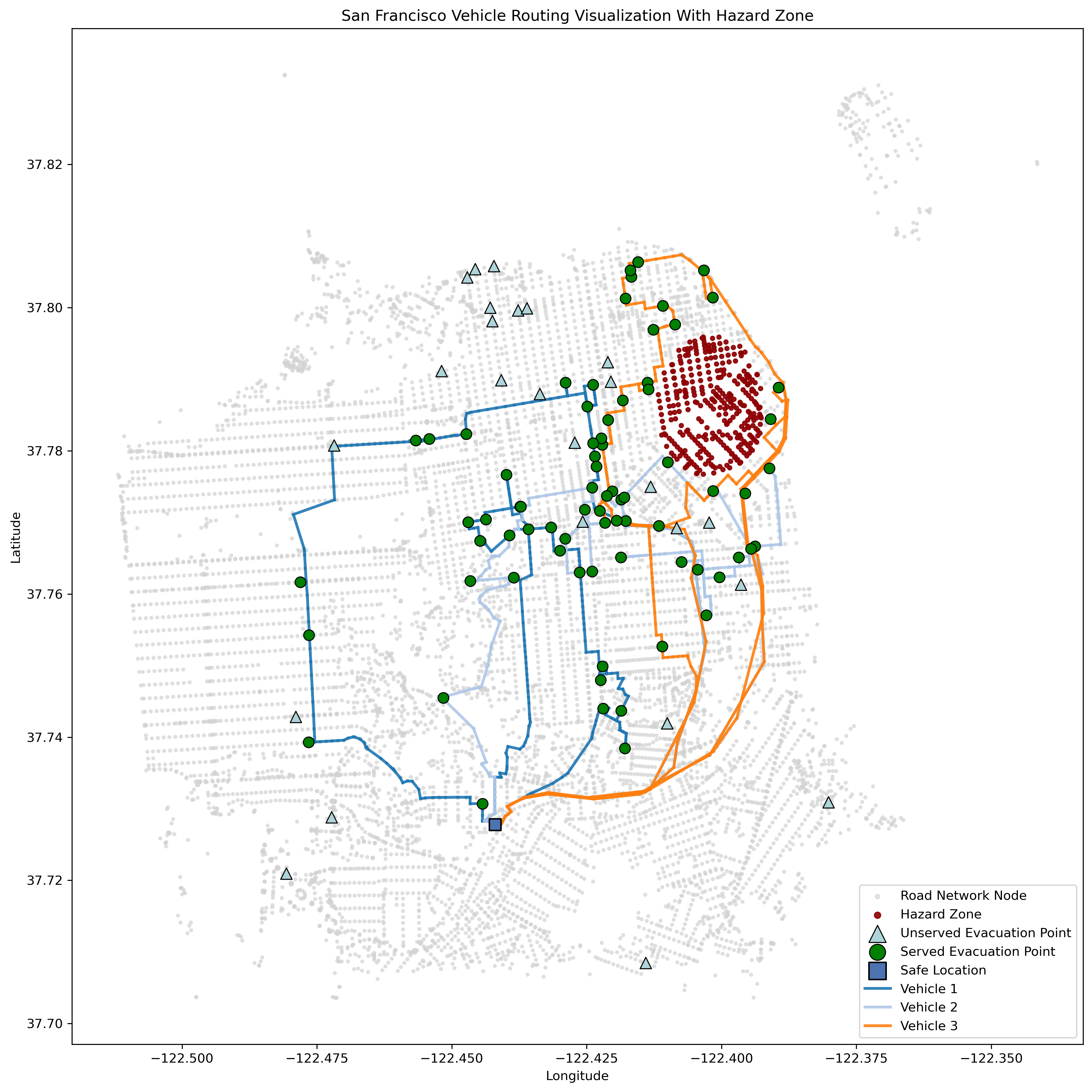}
    \label{fig:img312}
  }
  \hfill
  \subfloat[GREAT-EER solution ($99.67\%$ evacuation quota)]{
    \includegraphics[width=0.45\linewidth]{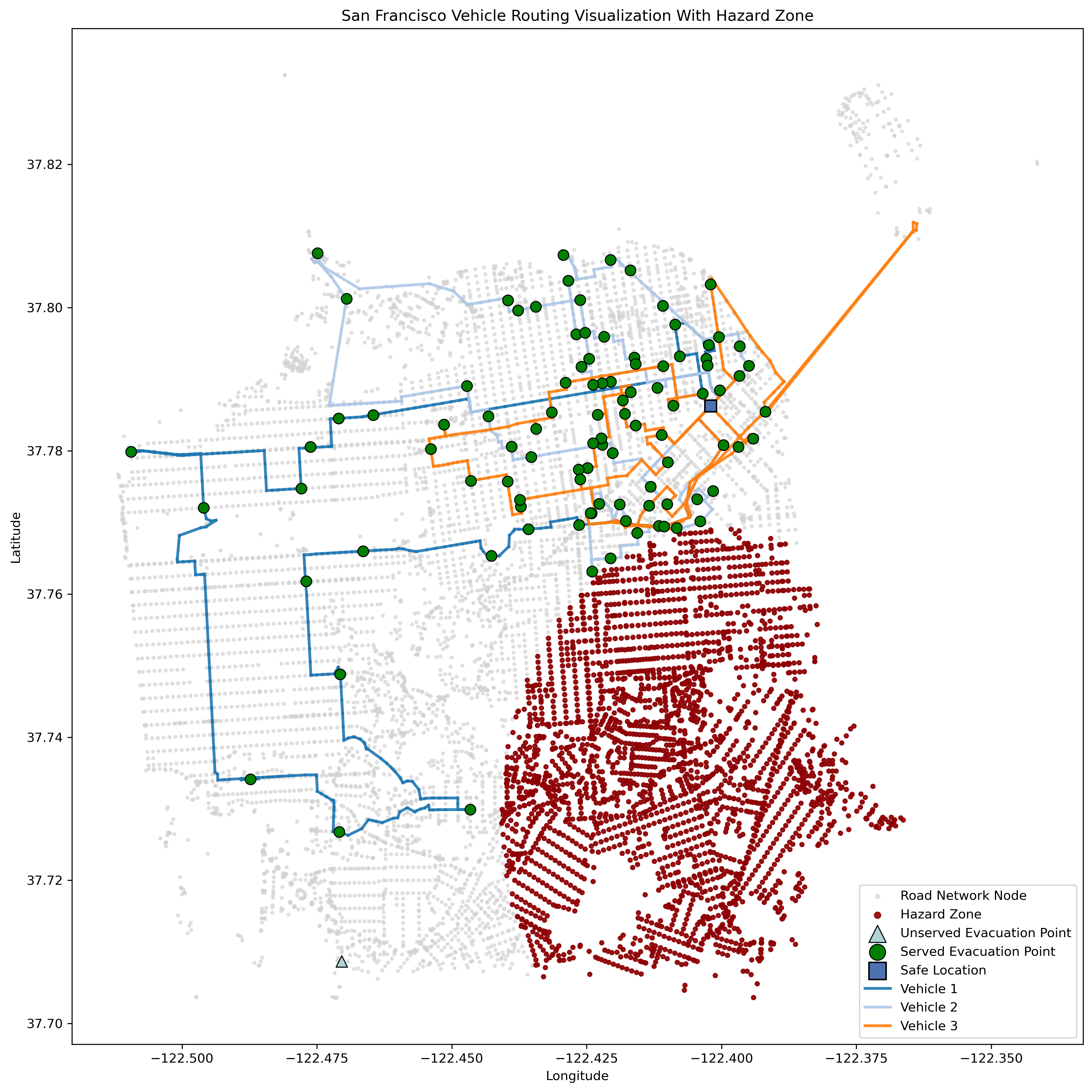}
    \label{fig:img212}
  }
  \hfill
  \subfloat[Greedy solution ($96.04\%$ evacuation quota)]{
    \includegraphics[width=0.45\linewidth]{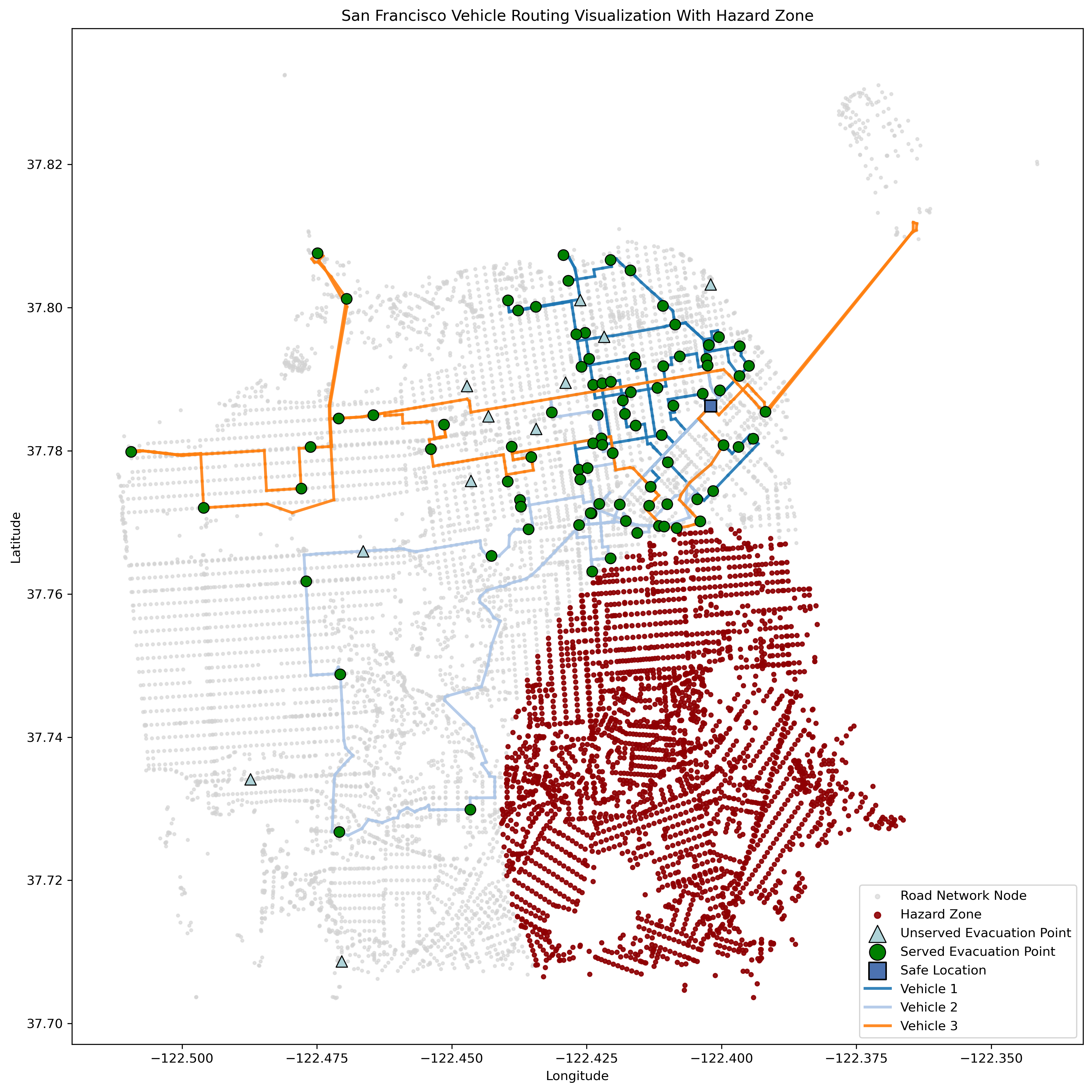}
    \label{fig:img412}
  }
  \caption{Example evacuations generated by GREAT-EER and greedy in San Francisco given 1.5h evacuation time with hazard zones}
  \label{fig2:exampe_evacuations}
\end{figure}

We provide two examples of this setting in Figure \ref{fig2:exampe_evacuations}. For both examples, we provide solutions generated by GREAT-EER and greedy side-by-side for comparison. 
An overview over the average evacuation quota (over 10 test instances) using GREAT-EER and the greedy baseline can be found in Table \ref{tab:great-vs-greedy-hazard}. We assume vehicles of capacity 50 and up to $30\%$ time window fraction.

In a similar out-of-distribution evaluation, we consider BEOP instances of 200 evacuation nodes, using 3-6 busses of capacity 50, $0.5$ to 2h of evacuation time and up to 30\% time window fraction. Note that these instances are much larger than the training instances (double the amount of evacuation nodes and up to double the amount of vehicles). The average performance (over 10 test instances) in terms of evacuation quota of GREAT-EER and the greedy baseline can be found in Table \ref{tab:great_vs_greedy_200}.

We can see that in both out-of-distribution cases (larger instances and hazard zones), GREAT-EER consistently outperforms the simple greedy baseline.

\begin{table}[t]
\centering
\caption{Average evacuation quota (over 10 instances) of GREAT-EER vs.\ greedy in the hazard zone setting. Higher is better.}
\label{tab:great-vs-greedy-hazard}
\begin{tabular}{c c | c c}
\toprule
\multicolumn{2}{c|}{Instance} & \multicolumn{2}{c}{Evacuation quota (\%)} \\
Vehicles & Time Limit & GREAT-EER & Greedy \\
\midrule
\multirow{3}{*}{1} & 0.5 & \textbf{33.03} & 24.64 \\
 & 1 & \textbf{56.61} & 45.18 \\
 & 1.5 & \textbf{72.33} & 61.37 \\
\midrule
\multirow{3}{*}{2} & 0.5 & \textbf{55.50} & 42.68 \\
 & 1 & \textbf{83.74} & 72.21 \\
 & 1.5 & \textbf{98.53} & 90.11 \\
\midrule
\multirow{3}{*}{3} & 0.5 & \textbf{70.75} & 58.62 \\
 & 1 & \textbf{97.99} & 90.13 \\
 & 1.5 & \textbf{100.0} & 99.37 \\
\midrule
\bottomrule
\end{tabular}
\end{table}

\begin{table}[t]
\centering
\caption{Average evacuation quota (over 10 instances) of GREAT-EER and greedy for instances of size 200. Higher is better.}
\label{tab:great_vs_greedy_200}
\begin{tabular}{cc|cc}
\toprule
\multicolumn{2}{c|}{Instance} & \multicolumn{2}{c}{Evacuation quota (\%)} \\
Vehicles & Time limit & GREAT-EER & Greedy \\
\midrule
\multirow{4}{*}{3} & 0.5 & \textbf{43.03} & 35.68 \\
 & 1 & \textbf{69.73} & 61.60 \\
 & 1.5 & \textbf{87.40} & 78.82 \\
 & 2 & \textbf{97.25} & 90.82 \\
\midrule
\multirow{4}{*}{4} & 0.5 & \textbf{52.35} & 44.10 \\
 & 1 & \textbf{81.31} & 73.09 \\
 & 1.5 & \textbf{97.07} & 90.83 \\
 & 2 & \textbf{99.97} & 98.38 \\
\midrule
\multirow{4}{*}{5} & 0.5 & \textbf{60.24} & 52.52 \\
 & 1 & \textbf{90.47} & 81.94 \\
 & 1.5 & \textbf{99.75} & 97.11 \\
 & 2 & \textbf{100.0} & 99.98 \\
\midrule
\multirow{4}{*}{6} & 0.5 & \textbf{66.92} & 59.99 \\
 & 1 & \textbf{96.22} & 89.39 \\
 & 1.5 & \textbf{100.0} & 99.53 \\
 & 2 & \textbf{100.0} & \textbf{100.0} \\
\midrule
\bottomrule
\end{tabular}
\end{table}

\subsubsection{Stochastic Online Evacuation}
So far, we have investigated how GREAT-EER can be used to generate deterministic evacuation plans in advance, before an emergency situation arises. 
However, during an emergency unexpected, stochastic situations can arise, e.g., travel times can be shorter or longer than expected (due to more or less traffic than expected) and people at evacuation points not waiting for the bus evacuation, but instead taking the car.
Therefore, we now compare the performance of deterministic evacuation plans generated by GREAT-EER before an emergency arises and a stochastic GREAT-EER model that dynamically reacts to changes in the environment. For example, the stochastic model can decide to visit a further evacuation node before returning to the depot because there are still available seats on the bus due to evacuees unexpectedly taking the car instead. Additionally, the stochastic model can react to the stochastic travel time nature and decide not to visit further evacuation points due to risk of violating the maximum evacuation time.

\begin{figure}
    \centering
    \includegraphics[width=0.5\linewidth]{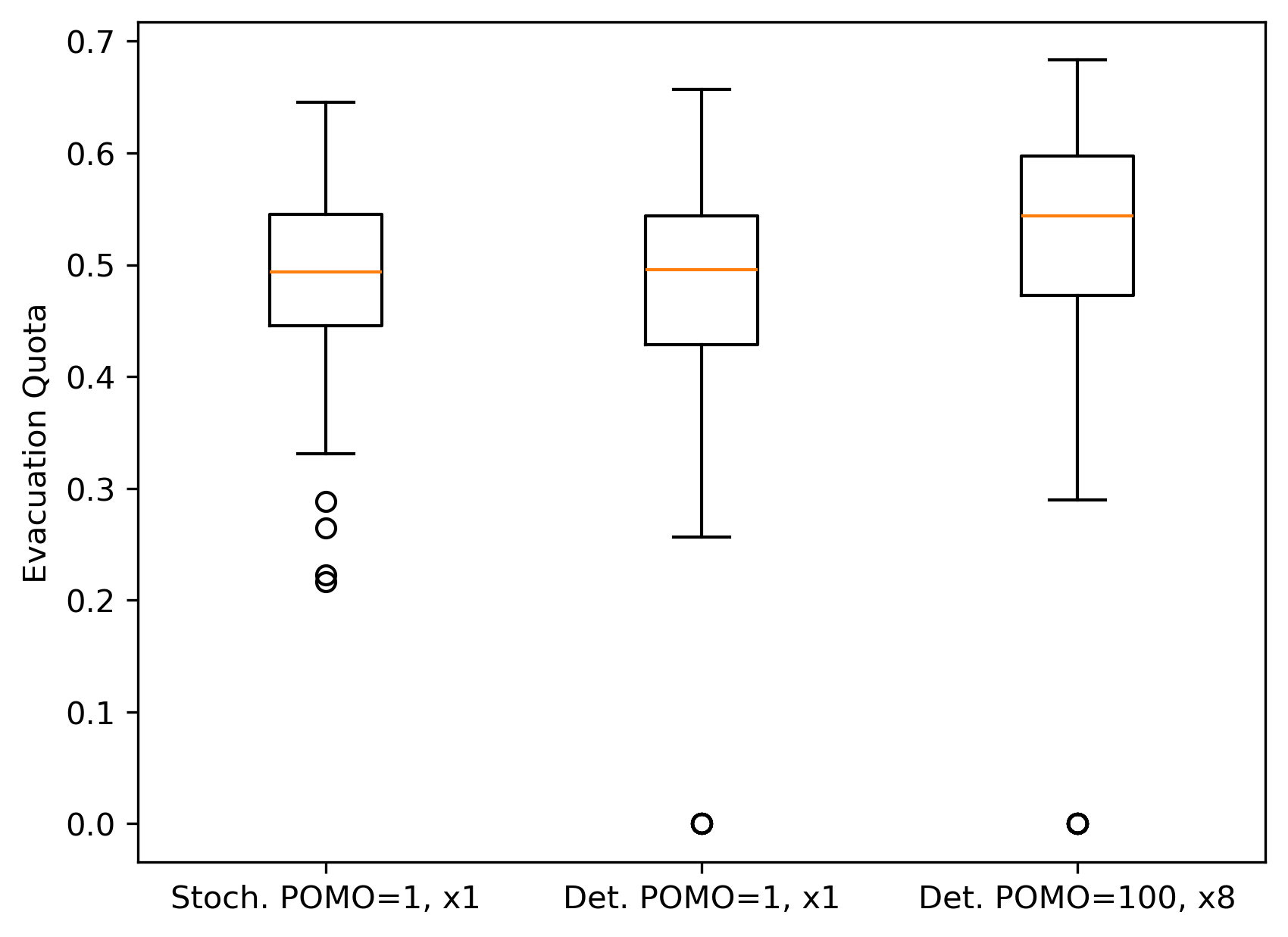}
    \caption{Evacuation quotas by stochastic GREAT-EER model and deterministic plans generated by GREAT-EER}
    \label{fig:stoch_vs_det}
\end{figure}

To demonstrate the advantage of using a stochastic model during real time evacuation over the deterministic model, we consider 100 test instances with 100 evacuation points using 1 vehicle with 50 seats during $1$h of evacuation time.
The resulting evacuation quotas on the 100 instances are shown in box plots in Figure \ref{fig:stoch_vs_det}.

We note that during deterministic planing we generally use instance augmentation and POMO rollouts to solve every instance multiple times, keeping the best solution the model generates. 
During stochastic, real-time operations this is, naturally, not possible, since the vehicle cannot try out different moves at once. Instead, we can only sample one solution trajectory. Therefore, in Figure \ref{fig:stoch_vs_det}, we report the performance of the stochastic model, the performance of the deterministic plan (using POMO=100 and $\times 8$ augmentation) and the performance of an additional deterministic plan that, similar to the stochastic model, does not do any augmentation.
If a model violates the maximum travel time for an instance, we set the corresponding evacuation quota to $0$ to represent an invalid evacuation.

In terms of evacuation quota, the performance of the stochastic model is roughly in between the augmentation-based deterministic plan and the deterministic plan without augmentation (when ignoring invalid plans).

More importantly, we note, that the stochastic model does not lead to any invalid evacuation scenarios, showing the advantage of the possibility to react to the stochastic realizations.
The deterministic planning leads to 15 and 9 invalid plans (out of 100) using augmentation and no augmentation, respectively.
We suspect that the increased amount of invalid plans when using the augmentation-based plan result from a tighter evacuation plan that uses the whole available travel time budget more effectively, leaving less slack. Due to the decreased slack, the plans are more prone to becoming invalid when travel times deviate from the expected values.

Further, we note that the stochastic model violates 0 time windows (note that generally due to unexpected increased travel time even the stochastic model can lead to time window violations). In comparison, the deterministic plans violate 4 (augmentation) and 2 (no augmentation) time windows. If such a time window violation occurs, we assume that the evacuees of such a node have already left thus not increasing the vehicle load but also not increasing the successful bus-evacuation quota while still having spent time on traveling to the node.

\section{Conclusion} \label{sec:conclusion}

In this paper we introduce an NP-hard combinatorial optimization problem that tackles emergency evacuations using buses called the bus evacuation orienteering problem (BEOP). The aim of this problem is to evacuate as many people as possible during a limited amount of time to reduce congestion occurring in car-based evacuation scenarios.
The problem is related to capacitated team orienteering problem with the key distinction of allowing several subtours for a vehicle. To the best of our knowledge, we are the first ones to consider this problem setting.
We model the BEOP as an MILP and a Markov Decision Process (MDP). 

We solve the BEOP using a deep reinforcement learning framework based on the graph edge attention network (GREAT), calling the overall framework graph edge attention network for emergency evacuation responses (GREAT-EER). 
To validate our model, we consider real-world data from San Francisco.
GREAT-EER is benchmarked against an exact solution method using an implementation of the MILP in an exact solver. By this, we can demonstrate that GREAT-EER achieves near optimal performance on small problem instances while being substantially faster. Further, solutions provided by GREAT-EER result in strong warm initializations for the exact solver in the MILP setting.
In larger evacuation scenarios, GREAT-EER is able to create solutions fast, while exact methods become prohibitively slow. The performance of GREAT-EER can be lower bounded by a simple greedy heuristic and we can show that GREAT-EER consistently outperforms such a simple heuristic on large BEOP instances.
We underline that GREAT-EER can be used by authorities for ex ante, provisional planning of evacuation plans during emergency situations. Further, GREAT-EER can used this way to determine how many vehicles of which capacities are needed given different evacuation times to achieve certain evacuation quotas.
We show that once trained GREAT-EER can also be applied to data distributions unseen during training, such as larger problem instances or instances with inaccessible hazard zones.
Additionally, we show that GREAT-EER can also be used in stochastic, real-time, online settings. By reacting dynamically to unforeseen changes during the evacuation scenario (changes in the number of evacuees or travel times due to increased traffic or blocked roads), GREAT-EER can prevent the construction of invalid solutions that violate hard constraints of the BEOP like the maximum evacuation time and lead to more effective usage of available vehicle capacities.

Beyond these experiments, we note that the extremely fast inference speed of GREAT-EER opens the possibility of additional use cases. For example, by solving each BEOP instance multiple times with different safe locations, we can select a spot that maximizes evacuation quota given a time limit, fixed bus parameters and vehicle number. Solving several BEOP instances at once is highly parallelizable when running GREAT-EER on GPUs (on a NVIDIA A40 GPU, solving 50 BEOP instances with 100 evacuation nodes in parallel takes 3 seconds). By this, it is possible to select a new safe locations within seconds in case parts of the city become inaccessible and existing plans become infeasible (compare hazard zone experiments).

In short, the contributions are:
\begin{itemize}
    \item Identification and introduction of the BEOP model for evacuations during emergency situations using busses.
    \begin{itemize}
        \item Formulation of an MILP for the BEOP.
        \item Formulation of an MDP for the BEOP.
    \end{itemize}
    \item Development of a reinforcement learning-based framework to solve the BEOP called GREAT-EER.
    \item Extensive real-world data-based experimental evaluation:
    \begin{itemize}
        \item Investigation of the performance of GREAT-EER in terms of solution quality and runtime compared to exact methods and heuristics.
        \item Investigation and validation of the performance of GREAT-EER in out-of-distribution settings.
        \item Ex ante, provisional evacuation planning using GREAT-EER on real-world settings using data from San Francisco, serving as guidelines for authorities to determine suitable evacuation resources.
        \item Stochastic evacuation operations using GREAT-EER in changing, dynamic evacuation scenarios with unforeseen changes to ensure valid evacuations and effective resource usage. 
    \end{itemize}
    
\end{itemize}

In future works, we aim to generalize our framework to further stochastic, real-time settings. In particular, we aim to consider several vehicles that cooperate in the stochastic setting which, in our MDP so far, is only possible in deterministic settings. Further, we want to consider more elaborate dynamic changes (congestion and travel times in dependence of evacuees choosing car-based evacuation instead of waiting for the bus), split-delivery and stochastic increased node demands or even stochastic new evacuation nodes. 
Additionally, there are several ways to extend our deterministic setting to be more realistic. Further works can extend the existing BEOP model with \textit{loading times} which reflect the time needed for evacuees entering the bus. While our existing model can easily extended to this setting, it was omitted in this work for the sake of simplicity. Moreover, it can be worth investigating settings with heterogeneous bus characteristics such as different passenger capacities or capacities dedicated to evacuees with special needs. Moreover, our model assumes the same reward for all evacuees. However, it could be possible to prioritize certain groups of evacuees, e.g., based on vulnerability. 
Further, we note that our solution approach is currently optimized for BEOP instances of 100 evacuation nodes. While experimental evaluation shows strong generalization performance to instances of 200 nodes, scaling to even larger instances is an open challenge and requires more advanced learning pipelines.

\section{Acknowledgments}

This work was performed as a part of the research project ``LEAR: Robust LEArning methods for electric vehicle Route selection'' funded by the Swedish Electromobility Centre (SEC). The computations were enabled by resources provided by the National Academic Infrastructure for Supercomputing in Sweden (NAISS) at Chalmers e-Commons partially funded by the Swedish Research Council through grant agreement no. 2022-06725.

\FloatBarrier

\printcredits

\bibliographystyle{cas-model2-names}

\bibliography{cas-refs}

\end{document}